\documentclass[a4paper]{article}

\usepackage[pages=all, color=black, position={current page.south}, placement=bottom, scale=1, opacity=1, vshift=5mm]{background}
\SetBgContents{
	\tt Preprint submitted to arXiv
}      

\usepackage[margin=1in]{geometry} 

\usepackage{amsmath}
\usepackage{amsthm}
\usepackage{amssymb}

\usepackage[utf8]{inputenc}
\usepackage{hyperref}
\hypersetup{hidelinks}

\usepackage{layouts}
\usepackage{xcolor}
\usepackage{pgf}
\usepackage{subcaption}
\usepackage{hhline}
\usepackage{array}
\newcolumntype{t}{!{\vrule width 2.5pt}}
\usepackage{makecell}
\usepackage{multirow}
\usepackage[super]{nth}
\usepackage{gensymb}
\usepackage{nicematrix}
\usepackage{float}

\usepackage[sort&compress,numbers,square]{natbib}
\theoremstyle{plain}

\theoremstyle{definition}

\usepackage{graphicx, color}

\usepackage{algorithm, algpseudocode}
\usepackage{mathrsfs}

\title{Inadequacy of common stochastic neural networks for reliable clinical decision support}
\author{Adrian Lindenmeyer$^{1,2,*}$, Malte Blattmann$^1$, Johannes Keller$^1$, Stefan Franke$^1$, \and Thomas Neumuth$^1$, Daniel Schneider$^1$}

\date{
\textit{$^1$Innovation Center Computer Assisted Surgery (ICCAS), Leipzig University, Semmelweisstrasse 14, Leipzig, 04103, Germany \\ 
$^2$Center for Scalable Data Analytics and Artificial Intelligence (ScaDS.AI) Dresden/Leipzig, Leipzig University, Germany} \\ [2ex]
\text{*adrian.lindenmeyer@iccas.de}\\[5ex]%
January 24, 2024
}

\begin{document}
\maketitle

\begin{abstract}
\textbf{Purpose:} 
Widespread adoption of AI for medical decision making is still hindered due to ethical and safety-related concerns.
For AI-based decision support systems in healthcare settings it is paramount to be reliable and trustworthy.
Common deep learning approaches, however, have the tendency towards overconfidence under data shift.
Such inappropriate extrapolation beyond evidence-based scenarios may have dire consequences. 
This highlights the importance of reliable estimation of local uncertainty and its communication to the end user.
While stochastic neural networks have been heralded as a potential solution to these issues, this study investigates their actual reliability in clinical applications.

\textbf{Methods:}
We centered our analysis on the exemplary use case of mortality/survival prediction for intensive care unit hospitalizations using electronic health records (EHR) from MIMIC3 study. 
For predictions on the EHR time series, Encoder-Only Transformer models were employed. 
Stochasticity of model functions was achieved  by incorporating common methods such as Bayesian neural network layers and model ensembles.

\textbf{Results:}
Our models achieve state of the art performance in terms of discrimination performance (AUC ROC: $0.868\pm0.011$, AUC PR: $0.554\pm0.034$) and calibration on the mortality prediction benchmark. However, epistemic uncertainty is critically underestimated by the selected stochastic deep learning methods. A heuristic proof for the responsible collapse of the posterior distribution is provided.

\textbf{Conclusions:} 
Our findings reveal the inadequacy of commonly used stochastic deep learning approaches to reliably recognize out-of-distribution (OoD) samples. 
In both methods, unsubstantiated model confidence is not prevented due to strongly biased functional posteriors, rendering them inappropriate for reliable clinical decision support.
This highlights the need for approaches with more strictly enforced or inherent distance-awareness to known data points, e.g., using kernel-based techniques.

\end{abstract}

\newpage
\section{Introduction}
The adoption of EHR in the healthcare sector promises remarkable potential for enhancing patient care and operational efficiency. 
These digital records encapsulate an immense volume of data, encompassing intricate patient histories, treatment pathways, diagnostic information, and clinical outcomes \citep{BenefitsAndDrawbacksOfElectronicHealthRecordSystems, BigDataAnalyticsInHealthcarePromiseAndPotential}. 
However, the sheer magnitude and complexity of EHR data presents a challenge by itself: it is beyond the capacity of human practitioners to effectively process and interpret this information in its entirety. 
This limitation necessitates the development of automated methods capable of discerning complex patterns, summarizing vast data \citep{BigDataAnalyticsInMedicineAndHealthcare, OmicAndElectronicHealthRecordBigDataAnalyticsForPrecisionMedicine, MachineLearningForPrecisionMedicine}, and indicating critical points that require human attention \citep{BuildingTrustInDeepLearningSystemTowardsAutomatedDiseaseDetection, AnalyzingTheRoleOfModelUncertaintyForElectronicHealthRecords, SecondOpinionNeededCommunicatingUncertaintyInMedicalMachineLearning}. 
Such advancements pave the way for sophisticated automated clinical decision support (CDS) systems, designed to augment the decision-making process in clinical settings \citep{MachineLearningForPrecisionMedicine}.

Machine learning (ML) and deep learning (DL) approaches have shown promising results in the analysis of EHR data in plenty of  studies centered around a multitude of predictive clinical applications \citep{DeepEHRASurveyOfRecentAdvancesInDeepLearningTechniquesForElectronicHealthRecordEHRAnalysis}.
In the healthcare domain, where decisions are usually safety-critical and ethically relevant, it is imperative that such automated methods employed in CDS systems are not only effective but also trustworthy \citep{BuildingTrustInDeepLearningSystemTowardsAutomatedDiseaseDetection, AnalyzingTheRoleOfModelUncertaintyForElectronicHealthRecords, TheNeedForUncertaintyQuantificationInMachineAssistedMedicalDecisionMaking, SecondOpinionNeededCommunicatingUncertaintyInMedicalMachineLearning, UncertaintyInDeepLearning}. 
While many studies certify striking predictive performance during model validation for in-distribution (ID) data, performance on OoD samples is naturally inconsistent \citep{AnalyzingTheRoleOfModelUncertaintyForElectronicHealthRecords}. 
Irrespective of the tradeoff between predictive capability and the risk of over-generalization, common ML and DL approaches strongly extrapolate from the available data \citep{UncertaintyInDeepLearning}.
Such excessive generalization may mislead end users into believing that there is substantial evidence supporting a prediction when, in reality, there may be none.
The consequence may be false decision making and decreased quality of medical care \citep{AReviewOfUncertaintyQuantificationInDeepLearningTechniquesApplicationsAndChallenges, TheNeedForUncertaintyQuantificationInMachineAssistedMedicalDecisionMaking, SecondOpinionNeededCommunicatingUncertaintyInMedicalMachineLearning, UncertaintyInDeepLearning}.
Distinguishing between complex ID and OoD data through human observation alone is likely unfeasible or completely impossible. 
While in most real-world applications deployed CDS systems are very likely to encounter OoD data, the resulting inability to decide which predictions to trust and which not renders the reliability of every prediction questionable.
Hence, it is crucial for automated CDS models to accurately convey the extent of evidence supporting their predictions, providing a measure of predictive certainty and thus, avoiding unsubstantiated assertions \citep{TheNeedForUncertaintyQuantificationInMachineAssistedMedicalDecisionMaking, AnalyzingTheRoleOfModelUncertaintyForElectronicHealthRecords, BuildingTrustInDeepLearningSystemTowardsAutomatedDiseaseDetection, SecondOpinionNeededCommunicatingUncertaintyInMedicalMachineLearning,UncertaintyInDeepLearning}.
Reliable estimation of predictive uncertainty enables collaborative frameworks \citep{MeasuringAndImprovingModelModeratorCollaborationUsingUncertaintyEstimation}.
In such schemes, the majority of clear-cut cases are efficiently managed by the automated methods, while ambiguous instances (OoD samples) are referred for review by a clinical professional.
This synergy ensures capitalization of the efficiency of automation while maintaining critical oversight of experienced medical practitioners.

Detection of OoD samples is intricately linked to accurate estimation of epistemic uncertainty. 
Epistemic or knowledge uncertainty represents the ambiguity in the model function learned from data. 
In general this ambiguity grows with increasing divergence from known data points (cf. OoD).
Unlike stochastic uncertainty, which stems from inherent data variability and is typically addressed by ML and DL models, epistemic uncertainty is unrecognized by such so-called point estimators \citep{UncertaintyInDeepLearning}.
In the quest to quantify epistemic uncertainty, common methodologies involve sampling from a functional posterior distribution that aligns with the training data \citep{AReviewOfUncertaintyQuantificationInDeepLearningTechniquesApplicationsAndChallenges}.
Among these methods, mean field approximation Bayesian neural networks and model ensembles stand out as particularly prominent approaches.
In this work, we critically evaluate the effectiveness of these models in estimating predictive uncertainty within a practical, application-driven CDS scenario. 
This evaluation is pivotal in understanding how these models perform in real-world healthcare settings, where the distinction between ID and OoD data is vital for making reliable clinical decisions.

In our work, we utilize data from the MIMIC3 study \citep{MIMIC3, MIMICIIIAFreelyAccessibleCriticalCareDatabase,PhysioBankPhysioToolkitAndPhysioNetComponentsOfANewResearchResourceForComplexPhysiologicSignals}, which comprises extensive EHR from intensive care unit (ICU) stays. 
To effectively process the time-dependent EHR data, we employ an Encoder-Only Transformer model (structurally derived from \citep{BEHRTTransformerforElectronic}). 
Transformers, known for their revolutionary impact in the field of natural language processing (NLP), are particularly adept at managing the sequential and contextual complexities inherent in EHR data. 
Our analysis is centered around binary outcome prognosis (survival prediction) for patients during their ICU stays. 
For this purpose, a benchmark dataset derived from the original MIMIC3 data was used \citep{HARUTYUNYANMultitaskLearningAndBenchmarkingWithClinicalTimeSeriesData}.
The analyzed use case is exemplary for data-driven CDS applications dealing with intricate EHR.

In the following, we provide a short overview of research related to our narrative. 
We then propose a data model designed to capture various clinical data modalities and unify them in a single embedding space for processing by the transformer models. 
Afterwards, we apply stochastic Encoder-Only architectures for survival classification of ICU timelines enabled to quantify predictive uncertainty. 
To our knowledge we are among the first to employ stochastic transformer-based architectures for applications in the medical domain.
Next, we show that our predictive models utilizing prominent stochastic deep learning approaches, while displaying striking performance on ID data, prove inadequate for accurate estimation of epistemic uncertainty on our example use case and may not be able to provide the reliability needed for safety-critical CDS applications.
Ultimately, we discuss reasons for their failure and promising novel advanced uncertainty quantification techniques from the literature.

\section{Related Work}

Early analysis of medical reasoning processes and the wish for computerised support in the medical field has been around since 1959 \citep{ReasoningFoundationsOfMedicalDiagnosis1959} and uncountable progress has been made in the field ever since \citep{MedBERTPretrainedContextualizedEmbeddingsOnLargescaleStructuredElectronicHealthRecordsForDiseasePrediction, DeepRepresentationLearningOfPatientDataFromElectronicHealthRecordsEHRASystematicReview, DeepEHRASurveyOfRecentAdvancesInDeepLearningTechniquesForElectronicHealthRecordEHRAnalysis}. While many different use cases such as image detection have received a lot of attention even in the medical field, the recent large scale collection of patient EHR data \citep{AdoptionOfElectronicHealthRecordsEHRsInChinaDuringThePast10YearsConsecutiveSurveyDataAnalysisAndComparisonOfSinoAmericanChallengesAndExperiences, AdoptionOfElectronicHealthRecordSystemsAmongUSNonFederalAcuteCareHospitals} and continued improvements in computational resources has lead to a whole new class of systems entering the medical field \citep{AttendAndDiagnoseClinicalTimeSeriesAnalysisUsingAttentionModels, DeepEHRASurveyOfRecentAdvancesInDeepLearningTechniquesForElectronicHealthRecordEHRAnalysis, PreTrainingOfGraphAugmentedTransformersForMedicationRecommendation}. Taping into the patients timeline these models access longitudinal information that has been known to clinicians to be of obvious importance for clinical decision making \citep{WhatCliniciansWantContextualizingExplainableMachineLearningForClinicalEndUse}. Models for EHR data need the ability to process sequential data. Its similarity to textual data has lead to the application of models known from the NLP domain, namely variants of RNNs and recently the Transformer \citep{AttentionIsAllYouNeed} and subvariant BERT \citep{BERT, DeepRepresentationLearningOfPatientDataFromElectronicHealthRecordsEHRASystematicReview}.

Recently \citep{DoctorAIPredictingClinicalEventsViaRecurrentNeuralNetworks} developed DoctorAI that predicts next visit diagnosis and medication codes utilizing a classical RNN based architecture. Later \citep{RETAINAnInterpretablePredictiveModelForHealthcareUsingReverseTimeAttentionMechanism} by the same group developed RETAIN, an architecture more specifically targeted and inspired by medical decision making. Two LSTM are used to direct attention at certain visits and features within those. Important information pieces are extracted, combined and passed through a classification MLP. \citep{HARUTYUNYANMultitaskLearningAndBenchmarkingWithClinicalTimeSeriesData} created a set of benchmarks based on the MIMIC3 data set \citep{MIMIC3} and rigorously tested different LSTM based architectures. To the authors knowledge these are the only rigorously defined benchmarks including preprocessing usable for objective comparison and was used in \citep{ModelingUncertaintyInDeepLearningModelsOfElectronicHealthRecords, ModelingTheUncertaintyInElectronicHealthRecordsABayesianDeepLearningApproach,  AttendAndDiagnoseClinicalTimeSeriesAnalysisUsingAttentionModels}. \citep{AttendAndDiagnoseClinicalTimeSeriesAnalysisUsingAttentionModels} apply a BERT architecture to the \citep{HARUTYUNYANMultitaskLearningAndBenchmarkingWithClinicalTimeSeriesData} benchmarks. Due to the Transformers inherent lack of recurrence, temporal information must be specifically given. In \citep{AttentionIsAllYouNeed} temporal information is given via temporal embedding and many others have relied on this approach. \citep{TemporalSelfAttentionNetworkforMedicalConceptEmbedding}, \citep{CEHRBERTIncorporatingTemporalInformationFrom} developed other explicit methods of including arbitrary temporal information. While \citep{TemporalSelfAttentionNetworkforMedicalConceptEmbedding} includes it in the attention layer by enriching the compatibility function of the attention block by the explicit temporal distance between two tokens, \citep{CEHRBERTIncorporatingTemporalInformationFrom} introduce time information as tokens into the transformer. \citep{PreTrainingOfGraphAugmentedTransformersForMedicationRecommendation} develop an embedding concept of medical tokens based on medical ontologies enriching information content of the tokens. Their model is a combination of a graph neural network for the embedding and a BERT-style Transformer for the inclusion of the longitudinal data aspect. \citep{MedBERTPretrainedContextualizedEmbeddingsOnLargescaleStructuredElectronicHealthRecordsForDiseasePrediction, BEHRTTransformerforElectronic, PredictingClinicalDiagnosisFromPatientsElectronicHealthRecordsUsingBERTbasedNeuralNetworks} utilize large real world medical data set showcasing the performance beyond typical benchmark data sets and potential real word applications. \citep{TAPERTimeAwarePatientEHRRepresentation} included information contained in medical notes by training two separate models and combining their embedding for various downstream tasks. \citep{Inpatient2VecMedicalRepresentationLearningForInpatients} utilize a Transformer architecture but focus specifically on semantical differences between in- and outpatients.

As seen in \citep{RETAINAnInterpretablePredictiveModelForHealthcareUsingReverseTimeAttentionMechanism} and \citep{ExplainablePredictionOfMedicalCodesFromClinicalText} additional targets such as explainability of the models decision are of high importance \citep{ModelingTheUncertaintyInElectronicHealthRecordsABayesianDeepLearningApproach, ModelingUncertaintyInDeepLearningModelsOfElectronicHealthRecords, AnalyzingTheRoleOfModelUncertaintyForElectronicHealthRecords}. This leads to the more recent focus on solving additional targets besides high accuracy for medical decision support. Captured by a recent study \citep{WhatCliniciansWantContextualizingExplainableMachineLearningForClinicalEndUse}, clinicians want a measure of uncertainty for a given decision. While uncertainty has been around as a general goal in AI for a while (a recent review was conducted by \citep{AReviewOfUncertaintyQuantificationInDeepLearningTechniquesApplicationsAndChallenges}) it has made its way into the medical space only recently as it is usually even more complex and costly to train. Nevertheless some papers have embraced its benefits and included uncertainty capabilities in their networks. 

\citep{ModelingTheUncertaintyInElectronicHealthRecordsABayesianDeepLearningApproach} learn heteroscedatic aleatoric uncertainty through optimisation/regularisation and later \citep{ModelingUncertaintyInDeepLearningModelsOfElectronicHealthRecords} investigate the capabilities of RNN and GRU based architectures to estimate different types of uncertainty. To produce epistemic uncertainties they use the dropout method and deep ensembles. Using uncertainty information they are able to significantly boost results when discarding uncertain examples and show negative correlations between certain data manipulations and resulting uncertainty. \citep{AnalyzingTheRoleOfModelUncertaintyForElectronicHealthRecords}
tested ensembles of LSTMs as well as different configurations of Bayesian LSTMs. They argue that capturing uncertainty is important for the identification and communication of cases where the models decision is likely to be questionable and more data should be collected.

\section{Methodology}

\subsection{Dataset} \label{sec:dataset}

MIMIC3 \citep{MIMIC3, MIMICIIIAFreelyAccessibleCriticalCareDatabase,PhysioBankPhysioToolkitAndPhysioNetComponentsOfANewResearchResourceForComplexPhysiologicSignals} is a large publicly available data set of roughly 40k patients who where admitted to the ICU at Beth Israel Deaconess Medical Center in Boston Massachusetts. It includes a wide range of clinically relevant information such as vital parameters, laboratory results, clinical procedures, medications, and outcome measures such as mortality. To date there are no established benchmark data sets for predictive tasks on EHR time series \citep{DeepEHRASurveyOfRecentAdvancesInDeepLearningTechniquesForElectronicHealthRecordEHRAnalysis}. Recently, a benchmark based on MIMIC3 was introduced in \citep{HARUTYUNYANMultitaskLearningAndBenchmarkingWithClinicalTimeSeriesData}. In their work, the authors delineate preprocessing methodologies, outline cohort selection strategies, identify multiple clinical predictive tasks, and set performance baselines for prevalent deep learning models.  Depending on the selected predictive task the cohort selection is slightly different to accommodate task specific exclusion criteria and the data is preprocessed according to the needed structure. For a more detailed description of the data and preprocessing steps see  \citep{HARUTYUNYANMultitaskLearningAndBenchmarkingWithClinicalTimeSeriesData}.

This work specifically uses the In-hospital mortality prediction task. At the core the data set is comprised of 21139 patients (further selected based on age, completeness of records, minimum length of stay, etc.) and a subset of 17 continuous and discrete features. Episodes begin at the time the patient is admitted to the ICU. Prediction of patient death occuring in-hospital is made 48 hours later. Signals are discretized to an hourly step size. A brief exploratory overview is given in table \ref{tab:dataset1}. 
As can be seen most continuous features have an average count per series of $\sim$42 showing their hourly measuring. On the other hand most categorical features relate to more global concepts such as Glasgow Come Scale (GCS) which is measured less frequently. We further group GCS into 5 bins as shown in table \ref{tab:dataset1}.

\begin{table}[]
\centering
\begin{NiceTabular}{| l | l | l | l l|}
\hline
\multicolumn{2}{| l |}{Feature} & \makecell{Avg. Count\\per Series} & \multicolumn{2}{| l |}{Distribution}\\
\hline \hline
\multicolumn{2}{| l |}{Heart rate [bpm]} & 43.97 & $\mu=86.26$ & $\sigma=17.88$\\
\hline
\multirow{3}{*}{\makecell[l]{BlooD\\preassure\\ [mmHg]}} & Diastolic & 42.88 & $\mu=60.49$ & $\sigma=14.30$\\
\cline{2-5}
 & Systolic & 42.89 & $\mu=120.25$ & $\sigma=22.14$\\
\cline{2-5}
 & Mean & 42.71 & $\mu=78.47$ & $\sigma=15.52$\\
\hline
\multicolumn{2}{| l |}{Fraction inspired oxygen [.]} & 2.95 & $\mu=0.54$ & $\sigma=0.19$\\
\hline
\multicolumn{2}{| l |}{Oxygen saturation [\%]} & 42.43 & $\mu=96.62$ & $\sigma=3.94$\\
\hline
\multicolumn{2}{| l |}{Respiratory rate [bpm]} & 43.28 & $\mu=19.30$ & $\sigma=6.06$\\
\hline
\multicolumn{2}{| l |}{Glucose [mg/dL]} & 12.37 & $\mu=143.08$ & $\sigma=64.42$\\
\hline
\multicolumn{2}{| l |}{pH [.]} & 5.98 & $\mu=7.36$ & $\sigma=0.12$\\
\hline
\multicolumn{2}{| l |}{Temperature [\degree C]} & 15.62 & $\mu=37.02$ & $\sigma=0.84$\\
\hline
\multicolumn{2}{| l |}{Height [cm]} & 0.19 & $\mu=168.59$ & $\sigma=13.90$\\
\hline
\multicolumn{2}{| l |}{Weight [kg]} & 1.48 & $\mu=83.12$ & $\sigma=24.25$\\

\hline \hline
\multicolumn{2}{| l |}{\multirow{2}{*}{Capillary refill rate} } & \multirow{2}{*}{0.17} & 13.43 \% & 1 \\

\multicolumn{2}{| l |}{\multirow{2}{*}{} } & & 86.57 \% & 0\\
\hline

\multirow{20}{*}{\makecell[l]{Glasgow \\ Coma \\ Scale \\ (GCS)}} & \multirow{4}{*}{Eye} & \multirow{4}{*}{14.83} & 57.30 \% & spontaneously (4) \\
 & & & 19.89 \% & to speech (3)\\
 & & & 6.48 \% & to pain(2)\\
 & & & 16.34 \% & no response (1)\\
\cline{2-5}

 & \multirow{5}{*}{Verbal} & \multirow{5}{*}{14.76} & 46.77 \% & oriented/normal (5)\\
 & & & 9.09 \% & confused (4)\\
 & & & 0.79 \% & inappropriate words (3)\\
 & & & 2.13 \% & incomprehensible (2)\\
 & & & 41.22 \% & no response (1)\\
\cline{2-5}

 & \multirow{6}{*}{Motor} & \multirow{6}{*}{14.76} & 69.90 \% & obeys commands (6)\\
 & & & 13.60 \% & localises pain (5)\\
 & & & 7.55 \% & flexion (4)\\
 & & & 0.77 \% & abnormal flexion (3)\\
 & & & 0.73 \% & abnormal extension (2)\\
 & & & 7.45 & no response (1)\%\\
\cline{2-5}

 & \multirow{5}{*}{Total} & \multirow{5}{*}{8.8} & 43.74 \% & 15 ("perfect")\\
 & & & 11.88 \% & 14-12 ("mild-moderate")\\
 & & & 20.86 \% & 11-9 ("moderate-severe")\\
 & & & 14.99 \% & 8-6 ("severe-coma")\\
 & & & 8.53 \% & 5-3 ("coma-brain dead")\\
\hline

\end{NiceTabular}
\caption{Summary of continuous and categorical features. Avg. Count per Series shows the average prevalence per patient. Distribution for continuous variables shows estimated mean ($\mu$) and standard deviation ($\sigma$). Distribution of categorical features is given by the percentage of observed categories.}
\label{tab:dataset1}
\end{table}

\subsection{Medical Datamodel for Transformer Application}\label{sec:datamodel}\label{sec:modelarchitecture}

EHR data can be viewed as a longitudinal stream of heterogeneous tokens. The token nomenclature is borrowed from the NLP domain and indeed there are similarities between language information and EHR records that have been exploited by numerous works \citep{CEHRBERTIncorporatingTemporalInformationFrom, AttendAndDiagnoseClinicalTimeSeriesAnalysisUsingAttentionModels, DeepEHRASurveyOfRecentAdvancesInDeepLearningTechniquesForElectronicHealthRecordEHRAnalysis, Inpatient2VecMedicalRepresentationLearningForInpatients, MedBERTPretrainedContextualizedEmbeddingsOnLargescaleStructuredElectronicHealthRecordsForDiseasePrediction}. There are, however, fundamental differences on the token level that need to be addressed \citep{MedBERTPretrainedContextualizedEmbeddingsOnLargescaleStructuredElectronicHealthRecordsForDiseasePrediction, TemporalSelfAttentionNetworkforMedicalConceptEmbedding}. "Medical tokens" are not restricted to words but encompass a multitude of heterogeneous concepts such as diagnostic values, laboratory results, vital parameters, medical imaging data, entire medical notes, medical procedures and medications, as well as data from the omics spectrum. Addtionally, medical records include a temporal component, in contrast to written language, which relies on information of order only.

For the task of outcome prognosis on EHR time series, we define two types of medical tokens, boolean ($\textit{token}^{(b)}$) and value tokens ($\textit{token}^{(v)}$) as shown in equations \ref{eq:btok} and \ref{eq:vtok}. Boolean tokens represent singular concepts (i.e. a patient's verbal response is categorized as \textit{confused} on the GCS). They also offer a method of conveying information of the absence of a concept (e.g., a patient does not suffer from diabetes). Without loss of generality this is not utilized as our data set does not supply information of i.e. diagnoses ruled out. Value tokens represent data points with an attached continuous value (e.g., measured heart rate of 88bpm). It extends the boolean token by a value. This allows it to represent values of zero (e.g., heart rate measured at 0bpm) distinguishable from its absence, as the boolean component conveys which concept the value is attached to. The timestamp is given by \(t\), \(c\) represents the concept and \(v\) represents a value. Further we define the cardinality of unique boolean and value tokens as $B$ and $V$.

\begin{align}
\textit{token}^{(b)}_{i} &= (t, c)_i \label{eq:btok} \\
\textit{token}^{(v)}_{i} &= (t, c, v)_i \label{eq:vtok}
\end{align}

The described tokenization formalism may readily be extended to vector valued concepts. This allows to represent data from entire images, graphs, or medical notes. To utilize these more complex data types for predictions, latent space representations can be used instead of the raw input data. These representations may be built with submodels each dedicated to a specific data type, which are trained simultaneously with the predictor. The latter is however not required for the present prediction task and is dropped for the remainder of this work.

Finally, each token is supplemented with a timestamp with a data set dependent temporal resolution. While a resolution of months or years may be adequate for EHR data involving medical history, a resolution of hours to minutes is needed for high-frequency data such as ICU records.

\begin{figure}
\centering
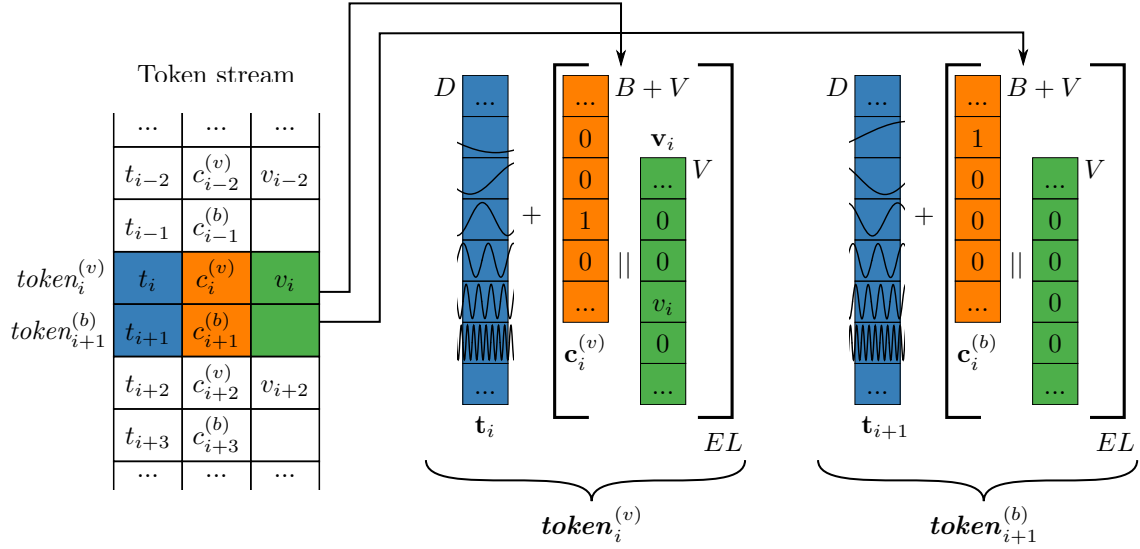
\caption{Visualisation of the embedding procedure.The parts of $\textit{token}^{(v/b)}$ tuples are encoded, embedded and combined. The concept part $c$ is encoded to a 1-hot vector $\boldsymbol{c}$. The value part $v$ is encoded to a 1-value vector $\boldsymbol{v}$. Both parts are concatenated ($||$) and embedded ($[.]_{EL}$). Linear time $t$ is embedded to circular time $\boldsymbol{t}$ via multiple sine/cosine waves of different frequencies. Time and concept embeddings are combined additively resulting in the final embedding $\textbf{\textit{token}}^{(v/b)}$.}
\label{fig:embedding}
\end{figure}

In this work we make use of an Encoder-Only Transformer architecture. The architecture remains unchanged except for the introduction of stochasticity for the quantification of uncertainty (see section \ref{sec:enabling_ue}). We modify the encoding and embedding procedure to ingest our previously defined data model (see figure \ref{fig:embedding}). The concepts \(c\) are one-hot encoded in a vector of length \(B + V\) (see figure \ref{fig:embedding} $\textit{\textbf{c}}^{(v/b)}$):

\begin{equation} \label{eq:concept_embedding}
\begin{aligned}
\boldsymbol{c}_{i} &= \{0,1\}^{B+V} \quad \text{and} \quad |\boldsymbol{c}_{i}| = 1
\end{aligned}
\end{equation}

Value tokens involve an additional encoding for the attached value. This is achieved by a derivative of the one-hot encoding previously shown. A one-hot vector of length \(V\) is multiplied with the corresponding value \(v\) to create a vector with zeros everywhere except for one entry which holds the value (see figure \ref{fig:embedding} $\textit{\textbf{v}}$):

\begin{equation} \label{eq:value_embedding}
\begin{aligned}
\boldsymbol{v}_{i} &= \{0,v_i\}^{V} \quad \text{and} \quad |\boldsymbol{v}_{i}| = v_i
\end{aligned}
\end{equation}

For the final encoding the two parts of a value tokens $\boldsymbol{c}$ and $\boldsymbol{v}$ are concatenated ($||$), while a boolean token is padded with zeros to the same size. The resulting encoding is passed through an embedding layer $EL: \mathbb{R}^{B+2V}\mapsto \mathbb{R}^{D}$ with $D$ being the chosen embedding size resulting in token specific embeddings (see figure \ref{fig:embedding} $[.]_{EL}$).

Time information $t$ is embedded directly to circular time of dimension $D$ utilizing the originally proposed approach by \citep{AttentionIsAllYouNeed}. The time embedding (see figure \ref{fig:embedding} $\boldsymbol{t}$) then results in the following vector which describes a geometric progression of sine and cosine waves of increasing wavelengths:

\begin{equation} \label{eq:circulartime}
\begin{aligned}
\boldsymbol{t}_{i} &= 
    \begin{cases}
          sin(t_i/10000^{2j/D}) & j \in 2k\\
          cos(t_i/10000^{2j/D}) & j \in 2k+1
    \end{cases} \\
    &j=1,...,D.
\end{aligned}
\end{equation}

Ultimately, time and concept embeddings are added resulting in the final embeddings (see figure \ref{fig:embedding} $\textbf{\textit{token}}^{(v/b)}$) shown in equation \ref{eq:final_embedding} and figure \ref{fig:embedding}.

\begin{equation} \label{eq:final_embedding}
\begin{aligned}
\textbf{\textit{token}}^{(b)}_{i} &= \textit{EL}(\boldsymbol{c}_{i}^{(b)} || \{0\}^{V})+\boldsymbol{t}_{i} \\
\textbf{\textit{token}}^{(v)}_{i} &= \textit{EL}(\boldsymbol{c}_{i}^{(v)} || \boldsymbol{v}_{i})+\boldsymbol{t}_{i}
\end{aligned}
\end{equation}

\subsection{Estimation of predictive Uncertainty}\label{sec:enabling_ue}

In the context of statistical inference, predictive uncertainty can be divided into aleatoric uncertainty (AU) describing stochastic uncertainty and epistemic uncertainty (EU) which stems from a lack of knowledge. 

AU is a limit of certainty for a given prediction that the model cannot surpass given the current feature set as data does not warrant less ambiguity. This can be due to stochastic behaviour of the modeled system (i.e. overlapping classes, measurement noise or inherent randomness in the outcome) and as such AU is proportional to these effects (i.e., high AU = large amount of noise).
EU is the ambiguity in the selection of the correct model function. Assuming certain smoothness constraints the set of possible functions within tightly sampled regions is small resulting in little relevant ambiguity in the model function. However in regions sparsely sampled or far away from known data points, the set of possible functions grows rapidly resulting in vast ambiguity about the model function. In the given context EU may be thought of as a measure of the predictions support by evidence \citep{AnalyzingTheRoleOfModelUncertaintyForElectronicHealthRecords, IntroductionToUncertaintyQuantification, UncertaintyQuantificationTheoryImplementationAndApplications}.

For a classification task neural networks are AU-aware by design as it is usually gained as a byproduct of the likelihood optimization \citep{ModelingUncertaintyInDeepLearningModelsOfElectronicHealthRecords, AnalyzingTheRoleOfModelUncertaintyForElectronicHealthRecords}), while EU is less easily accessible \citep{AnalyzingTheRoleOfModelUncertaintyForElectronicHealthRecords}. Hence attention is usually directed towards enabling a model to quantify EU. A recent paper by \citep{AReviewOfUncertaintyQuantificationInDeepLearningTechniquesApplicationsAndChallenges} reviewed a large number of current methods. In the scope of this paper we focus on two, a classical model ensemble approach and a Bayesian neural network with mean field approximation.

 \subsubsection{Ensemble Neural Networks}\label{sec:ENN}
 
An ensemble (ENN) consists of multiple individual models. Due to random initialization, stochastic optimization, high-dimensional loss surface and differing model behaviours, models build different solutions. The resulting difference in the built functions could theoretically approximate a posterior distribution from which EU can be measured. In areas of high sample density models will tightly agree on the solution thus signal low EU. Consequently in areas of low sample density or far outside the learned distribution, divergent model solutions should signal high EU \citep{AReviewOfUncertaintyQuantificationInDeepLearningTechniquesApplicationsAndChallenges}. Highly over-parameterised models such as neural networks should be especially suitable due to their universal function approximator capability that enables highly dissonant behaviours.

We utilize an approach similar to \citep{HyperparameterEnsemblesForRobustnessAndUncertaintyQuantification} making use of the models from a random parameter search. Of the $G$ models generated we select a subset $g \subset G$ of the best performing models for the ensemble. While some performance might be lost by this procedure, \citep{HyperparameterEnsemblesForRobustnessAndUncertaintyQuantification} argue for increased robustness and our results still show state-of-the-art performance while reducing computation time significantly.

\subsubsection{Bayesian Neural Networks}\label{sec:BNN}

The idea of Bayesian neural networks (BNN) have been extensively studied in \citep{APracticalBayesianFrameworkForBackpropagationNetworks, BayesianLearningForNeuralNetworks}. BNN place distributions over components of the model, e.g., the weights $p(w)$ of a neural network. The latter are updated based on Bayes rule and observed data points resulting in a posterior distribution. Bayesian networks promise to offer multiple benefits over classical non-Bayesian variants. Including a natural framework to quantify uncertainty and distinguish between AU and EU, Bayesian networks are less prone to overfitting due to their regularisation properties \citep{WeightUncertaintyInNeuralNetworks, HandsOnBayesianNeuralNetworksATutorialForDeepLearningUsers}. Generally the direct training of BNNs is intractable but multiple simplifying approaches exist. See \citep{AReviewOfUncertaintyQuantificationInDeepLearningTechniquesApplicationsAndChallenges, UncertaintyInDeepLearning} for an overview. A popular method is to use a variational inference approach with a mean field approximation in which the variational posterior $q(w)$ is simplified to a diagonal multivariate Normal $MVN(\boldsymbol{\mu}_q, I*\boldsymbol{\sigma}_q)$. Subsequently, Bayes by Backprop \citep{WeightUncertaintyInNeuralNetworks} adjusts the variational parameters $\theta_q = (\boldsymbol{\mu}_q, \boldsymbol{\sigma}_q)$ to minimize the evidence lower bound (ELBO) (equation \ref{eq:ELBO}). For practical applications the KL term may be adjusted by a factor $c_{kl}$ to determine the balance between these two parts. We use a radial posterior (equation \ref{eq:radial}) shown by \citep{RadialBayesianNeuralNetworksBeyondDiscreteSupportInLargeScaleBayesianDeepLearning} to improve results and increase training stability. For a detailed description of the used KL term see section \ref{sec:appx:KLTermderiv}.

\begin{equation} \label{eq:ELBO}
\begin{aligned}
ELBO &= c_{kl}*KL[q(w|\theta_q)||p(w)] -\mathbb{E}_{q(w|\theta_q)}[\log p(Y|w,X)]
\end{aligned}
\end{equation}

\subsection{Measuring Uncertainty}

Uncertainty can be measured in multiple different ways \citep{UncertaintyInDeepLearning}. For classification Mutual Information (MI) offers a well founded measure of disagreement between the model outputs. An approximation of $\text{MI}$ is defined through Predictive Entropy $\text{PE}$ by equation \ref{eq:mutualinformation} \citep{UncertaintyInDeepLearning}. Define the output logits as $\boldsymbol{\hat{l}} = \{\hat{l}_1, \hat{l}_2,...,\hat{l}_M\}$ with $M$ as the number of classes and $\mathcal{L} = \{\boldsymbol{\hat{l}}_{n}| n =1,...,N\}$ as the set of $N$ model evaluations for a single input. Other than Predictive Entropy, MI only reacts to differences in the returned distributions and not to the distributions itself. As such it is a measure of EU only. 

\begin{equation} \label{eq:mutualinformation}
\begin{aligned}
\boldsymbol{\hat{p}} &= Softmax(\boldsymbol{\hat{l}}) \\
\text{PE}(\mathcal{L}) &\approx -\sum_m^M \frac{1}{N}\sum_n^N\hat{p}_m\log\left(\frac{1}{N}\sum_n^N\hat{p}_m\right)\\
\mathbb{E}_{{\mathcal{L}}}[\text{PE}(\boldsymbol{\hat{l}})] &\approx \frac{1}{N}\sum_n^N\sum_m^M \hat{p}_m\log\left(\hat{p}_m\right)\\
\text{MI} &\approx \text{PE}(\mathcal{L}) - \mathbb{E}_{\mathcal{L}}[\text{PE}(\boldsymbol{\hat{l}})]
\end{aligned}
\end{equation}

A measure of similar properties can be constructed in logit space. Note that the Softmax function is invariant in the direction $n = norm((1,1,...,1)^{|M|})$. As a result the classification is related only to the perpendicular directions $\perp n$. We simply define the variance in these directions as a measure. While this might be a sophisticated task for higher dimensions, it simplifies in the two dimensional binary decision case. We define the signed distance between $n_{binary} = (1/\sqrt{2}, 1/\sqrt{2})$ and the logit $\boldsymbol{\hat{l}}$ as $\delta$ (see equation \ref{eq:delta}) and use the variance of $\delta$ as an uncertainty measure $\sigma(\delta)$.

\begin{equation} \label{eq:delta}
\begin{aligned}
\delta &= \boldsymbol{\hat{l}} \perp n \\
\end{aligned}
\end{equation}

Predictive Entropy averaged over model passes $\text{PE}(\mathcal{L})$ as defined in equation \ref{eq:mutualinformation} will later be used as a measure of uncertainty exclusively sensitive to AU.

\subsection{Model calibration}

We use matrix calibration \citep{BeyondTemperatureScalingObtainingWellCalibratedMultiClassProbabilitiesWithDirichletCalibration}. A linear transformation defined by a parameter matrix $\boldsymbol{W}^{MC}$ and bias $\boldsymbol{b}^{MC}$ is applied to the logit output $\boldsymbol{\hat{l}}$ of the model and trained to produce calibrated logits $\boldsymbol{\hat{l}}'$ (equation \ref{eq:matrixcalib}).

\begin{align} \label{eq:matrixcalib}
\boldsymbol{\hat{l}}' = \boldsymbol{W}^{MC}*\boldsymbol{\hat{l}} + \boldsymbol{b}^{MC}
\end{align}

Calibration is quantified by Expected Calibration Error (ECE) defined by the difference between accuracy and confidence (see equation \ref{eq:acc_conf}). With the use of binning scheme $\boldsymbol{K}$ a set of predictions $\boldsymbol{\hat{P}} = \{\boldsymbol{\hat{p}}_z\}$ are binned over predicted probabilities. The cardinality of bin $k$ is denoted by $K$ and refers to the number of predictions within bin $k$. $acc_{k,m}$ is the true proportion of class $m$ targets $y$ and $conf_{k,m}$ the average probability for class $m$ predictions within bin $k$. With slight abuse of notation, $z\in k$ denotes the sample indices $z$ for with $\hat{p}_{z,m}$ falls within the range of bin $k$.

\begin{align} \label{eq:acc_conf}
acc_{k,m} = \frac{1}{K}\sum_{z\in k}\boldsymbol{1}(y_z=m) \qquad conf_{k,m} = \frac{1}{K}\sum_{z\in k}\hat{p}_{z,m} 
\end{align}

By averaging ECE over a mini batch $\textit{s}$ we propose to use ECE directly as loss (see equation \ref{eq:eceloss_basic}). We denote $S$ as the cardinality of batch $s$. Again with slight abuse of notation, $z\in s$ denotes the sample indices $z$ within batch $s$ and $\boldsymbol{M}$ refers to the set of possible classes. Even though ECE involves discrete sorting operations modern fameworks enable the gradient flow over these operations. Nevertheless we found naive application of the ECE-loss to destabilise training especially in the case of ENN and applied further regularisation as described in \ref{sec:appx:eceregularistaion}.

\begin{align} \label{eq:eceloss_basic}
\textit{ECE-Loss} = \frac{1}{S*M}\sum_{z\in s}\sum_{m\in \boldsymbol{M}}\sum_{k\in \boldsymbol{K}} conf_{k,m} - acc_{k,m}
\end{align}

Calibration is not meant to further train its model. Because the calibration function consists of simply a linear matrix and bias we argue that this is satisfied. The method does not increase the models discriminatory power as it can only produce linear functions which are easily expressed by the network. Given common network sizes for modern applications the number of calibration parameters is negligible in comparison to the model parameters.

\section{Experiments} \label{sec:experiments}

We test two models based upon the described Transformer architecture (see section \ref{sec:modelarchitecture}). The Bayesian Transformer (BT) we give EU estimation capabilities by adding two bayesian layers (see section \ref{sec:BNN}) before the final layer outputting classes. By ensembling 33 distinct models, the Transformer Ensemble (TE) is equipped with ensemble EU estimation capabilities (see section \ref{sec:ENN}).
For each we calculate the three introduced measures of uncertainty, MI, $\sigma(\delta)$ and for comparison $\text{PE}(\mathcal{L})$, collectively referred to as $\text{m}_\text{U}$ for the remainder of this work.

We show performance for our models on the In-hospital mortality task from the \citep{HARUTYUNYANMultitaskLearningAndBenchmarkingWithClinicalTimeSeriesData} benchmarks (see section \ref{sec:dataset}) which uses 48 hours of patient intensive care signals and predicts if death occurs within the current hospital stay. For further details we refer the reader to the source publication.

We evaluate our two models against performance metrics (AUC ROC and AUC PR) of two models published in the benchmark \citep{HARUTYUNYANMultitaskLearningAndBenchmarkingWithClinicalTimeSeriesData}. The standard LSTM (S-LSTM), trained in a similar fashion to our two models on only the In-hospital mortaility use case. Secondly the multi task channel-wise LSTM Model (MTCW-LSTM) was selected as it represents the best model published with the benchmark while not being entirely comparable due to its multi task training.

To evaluate the reliability of $\text{m}_\text{U}$ returned by the models we order them in decreasing order and produce performance metrics while successively removing the most uncertain points down to the \nth{50} quantile in order to avoid sparse sampling artefacts. If the measures correlate with EU, one would expect the performance metrics to raise with increasingly restrictive quantile bounds. 

\begin{figure}
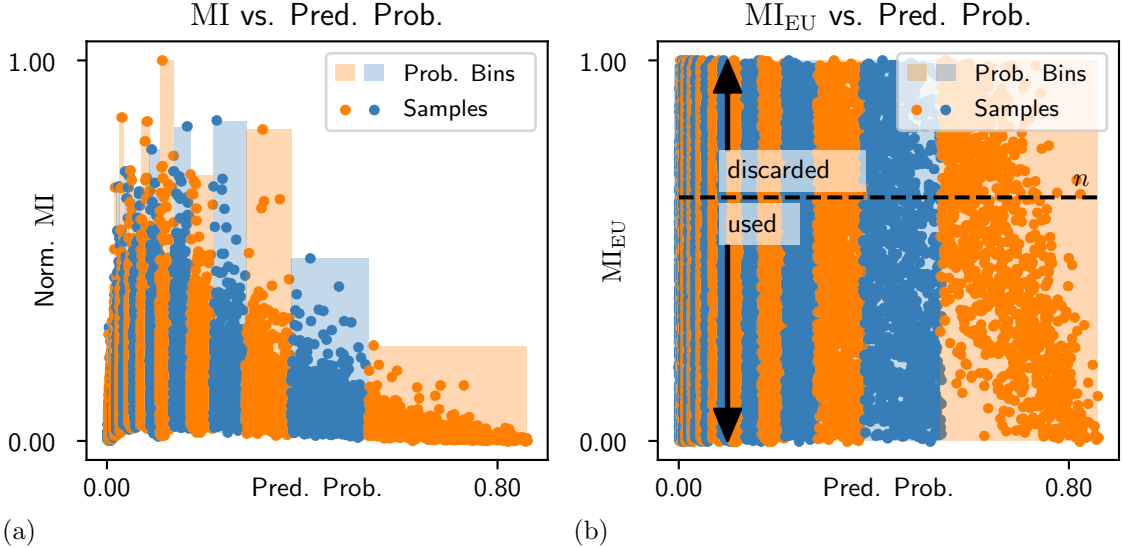

\centering
\captionsetup[subfigure]
{skip=-10pt,slc=off,margin={0pt,20pt}}
\subcaptionbox{\label{subfig:pre_remapping}}
{\input{remapping_pre_transformerensamble.pgf}}
\captionsetup[subfigure]
{skip=-10pt,slc=off,margin={0pt,20pt}}
\subcaptionbox{\label{subfig:post_remapping}}
{\input{remapping_post_transformerensamble.pgf}}
\captionsetup{subrefformat=parens}
\caption{Exemplary remapping for MI: \subref{subfig:pre_remapping} MI is binned by predicted probabilities. \subref{subfig:post_remapping} Within each bin the MI of the predictions is remapped to a uniform distribution. The successive cut-off procedure is indicated by the threshold $n$. By successively discarding predictions above the threshold $n$ only predictions with less model disagreement remain.}
\label{fig:remapping}
\end{figure}

Secondly we evaluate the epistemic $\text{m}_\text{U}$ (MI and $\sigma(\delta)$) on a more fundamental level by suppressing secondary correlations with the predicted probability (see figure \ref{fig:pvcs}) to study the EU quantification mechanisms in isolation. For any given probability, samples with higher $\text{m}_\text{U}$ values should have higher EU than samples with low $\text{m}_\text{U}$ values, resulting in less correct prediction for the former. 
To study the EU quantification mechanisms in isolation we separate predictions into bins according the probabilities and remap their $\text{m}_\text{U}$ values to a uniform distribution. The previously observed secondary correlations are almost completely removed resulting in remapped $\text{m}_\text{U}$
(see figure \ref{subfig:pre_remapping} for removal procedure and \ref{subfig:post_remapping} for remapped $\text{m}_\text{U}$). Again by gradually discarding the predictions with the largest $\text{m}_\text{U}$ (above a threshold $n$), rising performance metrics (i.e., AUC ROC) are an indication whether epistemic $\text{m}_\text{U}$ work as measure of EU in isolation.

\section{Results}

Discrimination performance results can be seen in figure \ref{fig:aucrocresults}. The best AUC ROC is achieved by the MTCW-LSTM from \citep{HARUTYUNYANMultitaskLearningAndBenchmarkingWithClinicalTimeSeriesData} with 0.870 closely followed by our TE with 0.868 which is within noise variance (\textpm $0.011$). Our BT performes sligthly worse at 0.862 than both TE and MTCW-LSTM while still showing better overall AUC ROC performance to their more comparable S-LSTM. Considering AUC PR our TE and BT models achieve superior results to their S-LSTM and MTCW-LSTM with an especially significant gap to the S-LSTM models showing much better performance concerning the minority class. The BT and MTCW-LSTM perform about on par within noise variance (\textpm 0.036) while the TE shows the overall best performance at 0.554.

The difference in AUC ROC ($\Delta$ AUC ROC) between the value calculated from all samples and the value calculated from the most certain \nth{50} quantiles by leveraging $\text{m}_{\text{U}}$ (both naive MI, $\text{PE}(\mathcal{L})$, $\sigma(\delta)$ and isolated MI$_{\text{EU}}$, $\sigma(\delta)_{\text{EU}}$) is also shown in figure \ref{fig:aucrocresults}. The full AUC ROC evolution over the range of quantiles, averaged over all splits, is shown in figures \ref{subfig:aucrocbt} and \ref{subfig:aucrocte}. Leveraging $\text{PE}(\mathcal{L})$ improves AUC ROC performance for both TE and BT by 0.075 and 0.073. MI also results in an increase in AUC ROC performance by about 0.056 and 0.037 for the TE and BT. However, once we isolate EU by stripping away the observed secondary correlations across the predicted probabilities, we can see that the benefit decreases significantly to only 0.014 and 0.002. $\sigma(\delta)$ turns out to have a disadvantageous negative effect on AUC ROC resulting in worse performance and a difference of -0.056 and -0.092. However remapping removes most of the negative correlation resulting in an insignificant delta in performance of 0.006 and -0.003. 

\begin{figure}
\centering
{\input{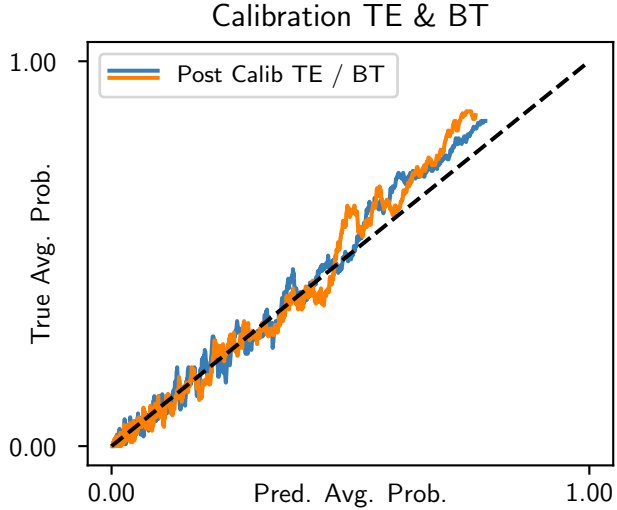}}
\caption{Calibration results for both TE and BT models. Calibration curves shows close to optimal behaviour (indicated by the diagonal) after calibration for both models.}
\label{fig:calibresults}
\end{figure}

Success of the calibration for both model types can be seen in figure \ref{fig:calibresults}. Calibration plots show probabilities well aligned with target frequencies close to optimal calibration as indicated by the diagonal.

The scores shown in figure \ref{fig:aucrocresults} are the result of a cross validation using 4 splits and swapping evaluation and test set for each split resulting in a total of 8 results for each model. The plots shown in the figures show only the mean results. Further details on the model selection and configurations are provided in \ref{sec:appx:modelselectionandconfigurations}.

\begin{figure}
\centering

\captionsetup[subfigure]
{skip=-10pt,slc=off,margin={0pt,20pt}}
\subcaptionbox{\label{subfig:aucrocbt}}
{\input{auc_roc_bayesiantransformer.pgf}}
\captionsetup[subfigure]
{skip=-10pt,slc=off,margin={0pt,20pt}}
\subcaptionbox{\label{subfig:aucrocte}}
{\input{auc_roc_transformerensamble.pgf}}
\captionsetup{subrefformat=parens}

\vspace{10pt}
{
\begin{tabular}{|l t c|c|c|c|}
\hline
 & TE & BT & S-LSTM & \makecell{MTCW-\\LSTM}\\
\hhline{|= t =|=|=|=|}
AUC ROC & 0.868 $\pm$0.011 & 0.862 $\pm$0.013 & \makecell[l]{0.855\\(0.835, 0.873)} & \makecell[l]{0.870\\(0.852, 0.887)} \\
\hline
AUC PR & 0.554 $\pm$0.034 & 0.546 $\pm$0.036 & \makecell[l]{0.485\\(0.431, 0.537)} & \makecell[l]{0.533\\(0.480, 0.584)} \\
\hhline{|= t =|=|=|=|}
\makecell[l]{$\Delta$ AUC ROC\\PE \nth{50}} & 0.075 $\pm$0.006 & 0.073 $\pm$0.006 & \textbf{------} & \textbf{------} \\
\hline
\makecell[l]{$\Delta$ AUC ROC\\MI \nth{50}} & 0.056 $\pm$0.010 & 0.037 $\pm$0.014 & \textbf{------} & \textbf{------} \\
\hline
\makecell[l]{$\Delta$ AUC ROC\\MI$_{\text{EU}}$ \nth{50}} & \textbf{0.014 $\pm$0.007} & \textbf{0.002 $\pm$0.007} & \textbf{------} & \textbf{------} \\
\hline
\makecell[l]{$\Delta$ AUC ROC\\$\sigma(\delta)$ \nth{50}} & -0.056 $\pm$0.017 & -0.092 $\pm$0.011 & \textbf{------} & \textbf{------} \\
\hline
\makecell[l]{$\Delta$ AUC ROC\\$\sigma(\delta)_{\text{EU}}$ \nth{50}} & \textbf{0.006 $\pm$0.008} & \textbf{-0.003 $\pm$0.015} & \textbf{------} & \textbf{------} \\
\hline

\end{tabular}
}

\caption{Correlations between discrimination performance and $\text{m}_{\text{U}}$. By the described procedure the most uncertain predictions, as signaled by naive MI, $\text{PE}(\mathcal{L})$, $\sigma(\delta)$ and isolated MI$_{\text{EU}}$ and $\sigma(\delta)_{\text{EU}}$, are successively discarded (Cut-off Quantile) for \subref{subfig:aucrocte} TE and \subref{subfig:aucrocbt} BT. The table shows the performance metrics for TE, BT and two benchmark models (S-LSTM and MTCW-LSTM) \citep{HARUTYUNYANMultitaskLearningAndBenchmarkingWithClinicalTimeSeriesData} ($\pm$ = 1 standard deviation, $(.,.)$ = 0.05/0.95 quantiles). The top two lines show discrimination performance on the entire data set in terms of AUC ROC and AUC PR. The lower portion shows the change in performance (+/-) when removing the 50\% most uncertain predictions. Scores shown in the table are the result of a cross validation using 4 splits and swapping evaluation and test set for each split resulting in a total of 8 results for each model. The plots shown in the figures show only the mean results.}
\label{fig:aucrocresults}
\end{figure}

\section{Discussion}

As seen in figures \ref{subfig:aucrocbt} and \ref{subfig:aucrocte} naive application of MI improves AUC ROC both for the BT and the TE while $\sigma(\delta)$ actually leads to a deterioration. We will show that this is not actually due to EU by the intended mechanics of model disagreement but rather due to favourable (or unfavourable) secondary correlations and its actual mode of action can be related to $\text{PE}(\mathcal{L})$ which only acts on AU.

\begin{figure}[ht]
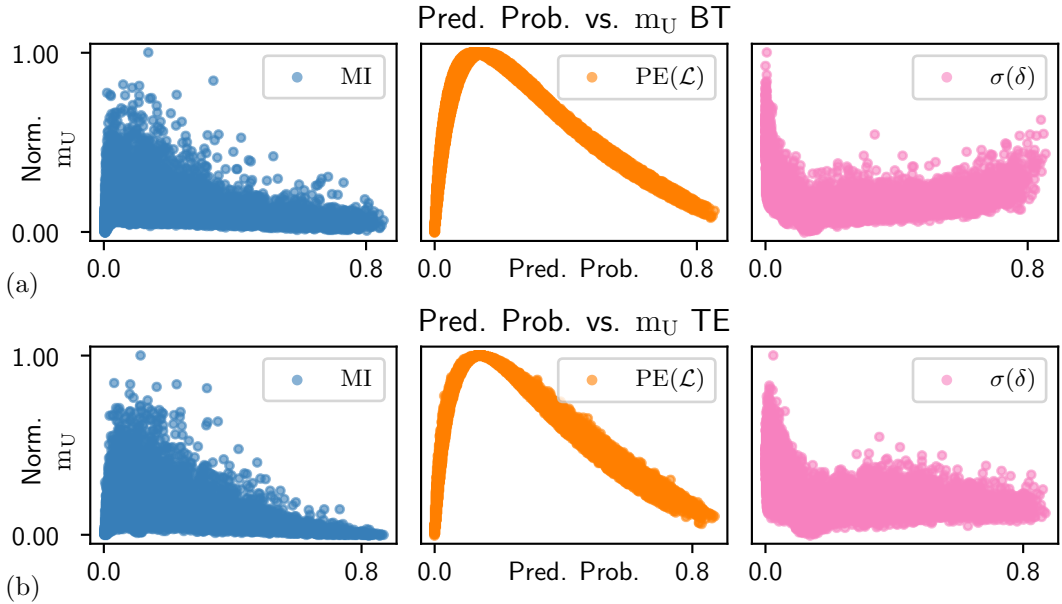

\centering
\captionsetup[subfigure]
{skip=-5pt,slc=off,margin={0pt,5pt}}
\subcaptionbox{\label{subfig:pvcsbt}}
{\input{pvcs_bayesiantransformer.pgf}}
\captionsetup[subfigure]
{skip=-5pt,slc=off,margin={0pt,5pt}}
\subcaptionbox{\label{subfig:pvcste}}
{\input{pvcs_transformerensamble.pgf}}
\captionsetup{subrefformat=parens}
\caption{Normed MI, $\sigma(\delta)$ and $\text{PE}(\mathcal{L})$ over pred. probabilities for \subref{subfig:pvcsbt} BT and \subref{subfig:pvcste} TE. Measures show secondary correlations evident by the correlations between pred. probabilities and the distribution of uncertainty. A correlation similar to $\text{PE}(\mathcal{L})$ (middle) is shown by MI and an inverted correlation is shown by $\sigma(\delta)$. These allow for an selection of samples statistically targeted at certain probabilities and as such AU.}
\label{fig:pvcs}
\end{figure}

Note first that ordering predicted samples by $\text{PE}(\mathcal{L})$ and then successively removing the most uncertain predictions leads to a selection towards extreme probabilities away from the prior (low AU predictions). In the binary case these are the classes 0 or 1 but this can be extended to any number of dimension. This can be seen in figure \ref{fig:pvcs}. This naturally leads to an increase in AUC ROC as it sorts out predictions close to the prior (high AU predictions) where the data, no matter how dense or sparse, is increasingly unclear and the classes overlap which trivially leads to a reduction in false predictions.

Observing the distribution of MI over the predictive probabilities in figure \ref{fig:pvcs} reveal that a similar ordering is achieved by MI which assigns higher uncertainty closer to the prior (high AU predictions). Thereby the positive correlation with AUC ROC can be attributed to the removal of predominately high AU predictions. The reason for MI diminishing towards extreme probabilities can be found in the Softmax function and its saturation at low entropy probabilities. Towards the extreme probabilities differences in the logits result in decreasing differences in the probits. In order to convey any meaningful disagreement registered by MI, networks would need to diverge much more strongly than observed during our test.

Counter intuitively the exact opposite is the case for $\sigma(\delta)$. While results do not diverge strong enough in logit space to register with MI due to the dampening effect of the Softmax function there is a correlation between low entropy predictions and increased divergence as can be seen in figure \ref{fig:pvcs}. This results in a reverse ordering to PE leading to the exclusion of samples at the extremes first naturally leading to an increase in false predictions.

While the previous paragraphs described the observed effects when naively employing MI or $\sigma(\delta)$ the remainder will focus on the inability of these measures to capture epistemic uncertainty reliably.

By isolating the EU mechanism (see section \ref{sec:experiments}) these secondary correlations across the probability spectrum and the ordering which this achieves are essentially removed leaving only the intended mechanism of model disagreement in place. As can be seen in figure \ref{fig:aucrocresults} this makes the uncertainty measures MI$_{\text{EU}}$, $\sigma(\delta)_{\text{EU}}$ almost completely inert. The invariance of AUC ROC measures mean MI$_{\text{EU}}$, $\sigma(\delta)_{\text{EU}}$ lead to a random selection of samples to discard. The intended mechanism of model disagreement behind both has no effect on unveiling uncertain samples.

The inability to properly model uncertainty when applying the mean field approximation to BNNs, and as such BT, is corroborated by \citep{UncertaintyInNeuralNetworksApproximatelyBayesianEnsembling, OnTheExpressivenessOfApproximateInferenceInBayesianNeuralNetworks}. However the inability of TE seemed a bit surprising at first as most literature suggests ENNs as the gold standard of uncertainty modelling.  Only \citep{SimpleAndPrincipledUncertaintyEstimationWithDeterministicDeepLearningViaDistanceAwareness} showed similar results with Deep Ensembles. 

Uncertainty by ensembles is based on the mechanism of random/dissonant behaviour outside data \citep{AReviewOfUncertaintyQuantificationInDeepLearningTechniquesApplicationsAndChallenges}. Modern sized neural networks certainly have the capacity to model a wide spectrum of behaviours given their universal function approximator property. For such mechanics the network would need to fundamentally "reserve some capacity" and spread it beneficially throughout the space in and around the data set. Without reserved capacity only the simplest linear functions between two spaces remain. Both requirements have no obvious reason to be satisfied and might not be the minimal loss solution and therefore the target of the training procedure. 

Modern neural networks mostly rely on ReLU activations due to their low overhead and currently Adam is seen as one of the obvious choices when training a neural network \citep{AnOverviewOfGradientDescentOptimizationAlgorithms}. The latter specifically encourages even weights with ultra sparse/low gradients to train by adapting the learning rate weight-wise \citep{AdamAMethodForStochasticOptimization}. The former leads to a response surface that is fundamentally piece-wise linear while the latter leads to a more global "network mobility" enabling the network to utilize its full potential to fit the given data. Lastly, weight decay, a method that penalizes the weights themselves is often used as one of many regularisation techniques.

In OoD regions, neural networks mostly exhibit linear behaviour when inter- or extrapolating. While our observations suggest that EU is severely underestimated in both cases, the underestimation of EU when interpolating might be more fundamental in nature and critical in application. When extrapolating tiny differences on the outside edges of the data set may lead to differences of the linear trajectories of each network. By extrapolating these diverge creating functional variance. There is however no control over this process and its effects might be of little use depending on the extend of the divergence. However, in the case of interpolation where the end points are given, reliable functional variance is even less likely to manifest. Again by considering the linearity and global mobility this means that all networks interpolate across the in-between space in a linear fashion. This results in virtually no increase in functional variance compared to ID data. 

\begin{figure}
\centering
{\input{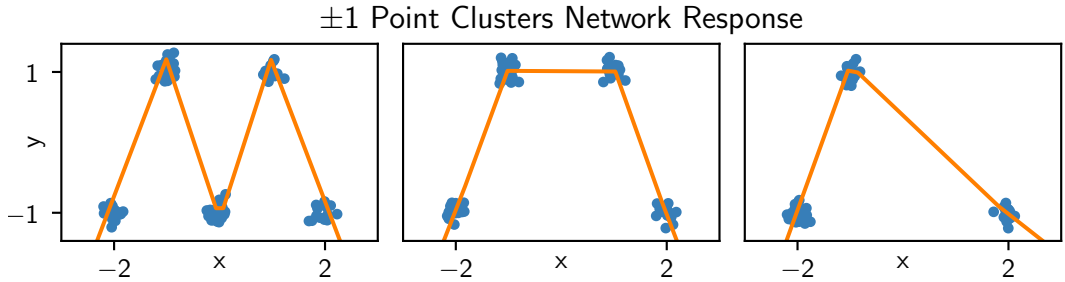}}
\caption{From left to right. Data set $\mathcal{D}$ (blue dots) with point clusters at $\pm1$ along the x-axis and the trained model function. From left to right, removal of selected point clusters results in collapse of the model function to the next simplest function.}
\label{fig:proof3}
\end{figure}

The collapse of functional variance during training can be shown empirically by a simple example shown in figure \ref{fig:proof3}. Assume a data set $\mathcal{D} = \{x_i, y_i\}$ (see equation \ref{eq:zigzagds}) resulting in small point clusters at $\pm1$ along the x-axis. During training randomly selected point clusters are removed from the data set. As can be seen in figure \ref{fig:proof3} the model collapses to the next simplest result by linear functions between the remaining point clusters. Model disagreement through random behavior is conditional on excessive non-linearity in sparse or OoD regions. Consequently for stable functional variance behaviour to form or at least hold during training the model would need to remain in previously found configurations.

\begin{equation} \label{eq:zigzagds}
\begin{aligned}
x_i &= j \pm\mathcal{N}(0, 0.1) \\
y_i &= (-1)^{j} \pm\mathcal{N}(0, 0.1) \\
    j &=-n,...,n.
\end{aligned}
\end{equation}

Hence neither ENNs nor BNNs are reliable methods for the modeling of epistemic uncertainty.

Shown results have far-reaching consequences for the application within medical or any other critical environment. Medically sufficient uncertainty recognition needs to go beyond the recognition of extrapolation uncertainty, i.e. when generated by false data entry resulting in negative blood pressures or similar, leading to embeddings far outside those previously trained on or supported by evidence. Which, as argued previously, might not be certain to work since there is no process in place to actually force functional variance to manifest. Medically sound uncertainty also needs to reliably recognize uncertainty within the "realistic simplex" i.e. patients with realistic features but in a previously unknown combination and as such not well supported by evidence, without being at the mercy of favorable training dynamics and the hopes of conserved functional variance in the case of ENN. The opinion of the authors of this work is that the latter type of uncertainty, which neither BNN nor ENN could reliably provide, might be the one of even greater medical significance as it is related to the amount of knowledge behind the models prediction and less to a general difference between the data the model was trained on and the data it is applied to.

\section{Conclusion}

In this study, we explored the feasibility of integrating transformer architectures with built-in stochastic components for reliable applications in CDS, focusing on the task of ICU survival prediction. 
For this purpose, we developed a data model to process multimodal time-dependent EHR and combined Encoder-Only Transformer architectures with common stochastic methods for epistemic uncertainty quantification including mean field Bayesian neural network layers and a model ensemble approach.
We applied these models on a benchmark task involving data from the MIMIC3 study.
Our models demonstrated very convincing predictive performance within the data set during cross-validation. 
However, the primary objective of this study was to assess whether the incorporated stochastic methods could enhance the reliability and trustworthiness of the transformer models, representing critical requirements in safety-sensitive medical decision-making.
Unfortunately, our findings indicate that both methods under investigation - mean field Bayesian layers and ensemble approach - fall short in adequately estimating epistemic uncertainty and consequently, do not meet the criteria necessary for ensuring that predictions are based on sufficient evidence. 
We observed that introduced variability by stochastic weights (Bayesian layers) or randomized initial parameter configurations (ensemble) does not necessarily translate into variability in the sampled model functions after optimization. 
This results in models producing heavily biased functional posteriors with insufficient functional variance in regions far away from the training data.
The models used in this study focused their entire capacity on modeling ID data during optimization, leading to an oversimplification of functional variance in the OoD regions. 
Evidence from our work suggests that even more sophisticated ensemble methods, such as ensembles containing multiple different DL architectures, hypermodels \citep{HypermodelsForExploration}, deep ensembles \citep{SimpleAndScalablePredictiveUncertaintyEstimationUsingDeepEnsembles}, and Stein ensembles \citep{OnSteinVariationalNeuralNetworkEnsembles}, may not generate appropriate functional posteriors as unbiased as expected.
Their inability to reliably detect OoD samples limits their utility in collaborative CDS frameworks and medical decision making as a whole.
These observations point to the necessity for alternative approaches where the variability of posterior functions is specifically enforced and optimized in balance with predictive performance. 
Promising techniques involve kernel-based solutions, which incorporate means of measuring the distance to known data. 
Such methods include e.g., Deep Gaussian processes \citep{DeepBayesianGaussianProcessesForUncertaintyEstimationInElectronicHealthRecords} and Spectral Normalised Neural Gaussian Processes \citep{SimpleAndPrincipledUncertaintyEstimationWithDeterministicDeepLearningViaDistanceAwareness}.
Further investigation into the epistemic uncertainty quantification capacities of these alternative methods, especially when applied to complex EHR data for CDS, remains an area for future research. 
This continued exploration is vital for developing more robust and reliable CDS systems capable of assessing the limits of their given evidence and enabling widespread adoption of DL in critical medical decision-making processes.

\paragraph{Acknowledgements} \mbox{} \\
A.L. acknowledges the financial support by the Federal Ministry of Education and Research of Germany and by Saechsisches Staatsministerium für Wissenschaft, Kultur und Tourismus in the programme Center of Excellence for AI-research "Center for Scalable Data Analytics and Artificial Intelligence Dresden/Leipzig“, project identification number: ScaDS.AI. M.B., S.F., T.N. and D.S. acknowledge funding by the Deutsche Forschungsgemeinschaft (DFG, German Research Foundation) – 460234259 (NFDI/34/1).

\noindent The dataset supporting the conclusions of this article is the Medical Information Mart for Intensive Care version 3 (MIMIC3). More details are available at https://physionet.org/content/mimiciii/1.4/. Our access to the database was approved after completion of the collaborative institutional training initiative (CITI program) web-based training course, “Data or Specimens Only research” (Record ID:42586885).

\paragraph{Data availability} \mbox{} \\
The MIMIC3 database is available at https://physionet.org/content/mimiciii/1.4/.

\paragraph{Code availability} \mbox{} \\
Code files (including model definitions, training procedure and evaluation metrics) will be uploaded to a git repository to which access will be granted upon request.
	
\newpage
\bibliography{refs}

\begin{thebibliography}{57}
\providecommand{\natexlab}[1]{#1}
\providecommand{\url}[1]{\texttt{#1}}
\expandafter\ifx\csname urlstyle\endcsname\relax
  \providecommand{\doi}[1]{doi: #1}\else
  \providecommand{\doi}{doi: \begingroup \urlstyle{rm}\Url}\fi

\bibitem[Abdar et~al.(2021)Abdar, Pourpanah, Hussain, Rezazadegan, Liu, Ghavamzadeh, Fieguth, Cao, Khosravi, Acharya, Makarenkov, and Nahavandi]{AReviewOfUncertaintyQuantificationInDeepLearningTechniquesApplicationsAndChallenges}
M.~Abdar, F.~Pourpanah, S.~Hussain, D.~Rezazadegan, Li~Liu, M.~Ghavamzadeh, P.~Fieguth, Xiaochun Cao, A.~Khosravi, U.~R. Acharya, V.~Makarenkov, and S.~Nahavandi.
\newblock A review of uncertainty quantification in deep learning: Techniques, applications and challenges.
\newblock \emph{Information Fusion}, 76:\penalty0 243--297, 12 2021.
\newblock \href{https://doi.org/10.1016/j.inffus.2021.05.008}{\ttfamily\path{ doi:10.1016/j.inffus.2021.05.008}}.

\bibitem[Begoli et~al.(2019)Begoli, Bhattacharya, and Kusnezov]{TheNeedForUncertaintyQuantificationInMachineAssistedMedicalDecisionMaking}
E.~Begoli, T.~Bhattacharya, and D.~Kusnezov.
\newblock The need for uncertainty quantification in machine-assisted medical decision making.
\newblock \emph{Nature Machine Intelligence}, 1:\penalty0 20--23, 1 2019.
\newblock \href{https://doi.org/10.1038/s42256-018-0004-1}{\ttfamily\path{ doi:10.1038/s42256-018-0004-1}}.

\bibitem[Blinov et~al.(2020)Blinov, Avetisian, Kokh, Umerenkov, and Tuzhilin]{PredictingClinicalDiagnosisFromPatientsElectronicHealthRecordsUsingBERTbasedNeuralNetworks}
P.~Blinov, M.~Avetisian, V.~Kokh, D.~Umerenkov, and A.~Tuzhilin.
\newblock Predicting clinical diagnosis from patients electronic health records using bert-based neural networks.
\newblock In \emph{Artificial Intelligence in Medicine: 18th International Conference on Artificial Intelligence in Medicine}, pages 111--121, 2020.
\newblock \href{https://doi.org/10.1007/978-3-030-59137-3-11}{\ttfamily\path{ doi:10.1007/978-3-030-59137-3-11}}.

\bibitem[Blundell et~al.(2015)Blundell, Cornebise, Kavukcuoglu, and Wierstra]{WeightUncertaintyInNeuralNetworks}
C.~Blundell, J.~Cornebise, K.~Kavukcuoglu, and D.~Wierstra.
\newblock Weight uncertainty in neural networks.
\newblock In \emph{Proceedings of the 32nd International Conference on Machine Learning}, volume~37, pages 1613--1622, 07 2015.

\bibitem[Choi et~al.(2016{\natexlab{a}})Choi, Bahadori, Kulas, Schuetz, Stewart, and Sun]{RETAINAnInterpretablePredictiveModelForHealthcareUsingReverseTimeAttentionMechanism}
Edward Choi, M.~T. Bahadori, J.~A. Kulas, A.~Schuetz, W.~F. Stewart, and Jinmeng Sun.
\newblock Retain: An interpretable predictive model for healthcare using reverse time attention mechanism.
\newblock In \emph{Proceedings of the 30th International Conference on Neural Information Processing Systems}, page 3512–3520, 12 2016{\natexlab{a}}.

\bibitem[Choi et~al.(2016{\natexlab{b}})Choi, Bahadori, Schuetz, Stewart, and Sun]{DoctorAIPredictingClinicalEventsViaRecurrentNeuralNetworks}
Edward Choi, M.~T. Bahadori, A.~Schuetz, W.~F. Stewart, and Jimeng Sun.
\newblock Doctor ai: Predicting clinical events via recurrent neural networks.
\newblock In \emph{Proceedings of the 1st Machine Learning for Healthcare Conference}, volume~56, pages 301--318, 08 2016{\natexlab{b}}.

\bibitem[D'Angelo et~al.(2021)D'Angelo, Fortuin, and Wenzel]{OnSteinVariationalNeuralNetworkEnsembles}
F.~D'Angelo, V.~Fortuin, and F.~Wenzel.
\newblock On stein variational neural network ensembles.
\newblock 2021.
\newblock \href{http://arxiv.org/abs/2106.10760}{{\ttfamily arXiv:2106.10760 [cs.LG]}}.

\bibitem[Darabi et~al.(2020)Darabi, Kachuee, Fazeli, and Sarrafzadeh]{TAPERTimeAwarePatientEHRRepresentation}
S.~Darabi, M.~Kachuee, S.~Fazeli, and M.~Sarrafzadeh.
\newblock Taper: Time-aware patient ehr representation.
\newblock \emph{IEEE Journal of Biomedical and Health Informatics}, 24\penalty0 (11):\penalty0 3268--3275, 11 2020.
\newblock \href{https://doi.org/10.1109/jbhi.2020.2984931}{\ttfamily\path{ doi:10.1109/jbhi.2020.2984931}}.

\bibitem[Devlin et~al.(2019)Devlin, Chang, Lee, and Toutanova]{BERT}
J.~Devlin, {Ming-Wei} Chang, Kenton Lee, and K.~Toutanova.
\newblock Bert: Pre-training of deep bidirectional transformers for language understanding.
\newblock In \emph{Proceedings of the 2019 Conference of the North American Chapter of the Association for Computational Linguistics: Human Language Technologies}, volume~1, pages 4171--4186, 06 2019.
\newblock \href{https://doi.org/10.18653/v1/n19-1423}{\ttfamily\path{ doi:10.18653/v1/n19-1423}}.

\bibitem[Dusenberry et~al.(2020)Dusenberry, Tran, Choi, Kemp, Nixon, Jerfel, Heller, and Dai]{AnalyzingTheRoleOfModelUncertaintyForElectronicHealthRecords}
M.~W. Dusenberry, Dustin Tran, Edward Choi, J.~Kemp, J.~Nixon, G.~Jerfel, K.~Heller, and Andrew~M. Dai.
\newblock Analyzing the role of model uncertainty for electronic health records.
\newblock In \emph{Proceedings of the ACM Conference on Health, Inference, and Learning}, pages 204--213, 04 2020.
\newblock \href{https://doi.org/10.1145/3368555.3384457}{\ttfamily\path{ doi:10.1145/3368555.3384457}}.

\bibitem[Dwaracherla et~al.(2020)Dwaracherla, Lu, Ibrahimi, Osband, Wen, and Roy]{HypermodelsForExploration}
V.~Dwaracherla, Xiuyuan Lu, M.~Ibrahimi, I.~Osband, Zheng Wen, and B.~Van Roy.
\newblock Hypermodels for exploration.
\newblock 2020.
\newblock \href{http://arxiv.org/abs/2006.07464}{{\ttfamily arXiv:2006.07464 [cs.LG]}}.

\bibitem[Farquhar et~al.(2020)Farquhar, Osborne, and Gal]{RadialBayesianNeuralNetworksBeyondDiscreteSupportInLargeScaleBayesianDeepLearning}
S.~Farquhar, M.~A. Osborne, and Y.~Gal.
\newblock Radial bayesian neural networks: Beyond discrete support in large-scale bayesian deep learning.
\newblock In \emph{Proceedings of the Twenty Third International Conference on Artificial Intelligence and Statistics}, volume 108, pages 1352--1362, 08 2020.

\bibitem[Foong et~al.(2020)Foong, Burt, Li, and Turner]{OnTheExpressivenessOfApproximateInferenceInBayesianNeuralNetworks}
Andrew Foong, D.~Burt, Yingzhen Li, and R.~Turner.
\newblock On the expressiveness of approximate inference in bayesian neural networks.
\newblock In \emph{Advances in Neural Information Processing Systems}, volume~33, pages 15897--15908, 2020.

\bibitem[Gal(2016)]{UncertaintyInDeepLearning}
Y.~Gal.
\newblock \emph{Uncertainty in Deep Learning}.
\newblock PhD thesis, University of Cambridge, 2016.

\bibitem[Goldberger et~al.(2000)Goldberger, Amaral, Glass, Hausdorff, Ivanov, Mark, Mietus, Moody, Peng, and Stanley]{PhysioBankPhysioToolkitAndPhysioNetComponentsOfANewResearchResourceForComplexPhysiologicSignals}
A.~L. Goldberger, L.~A. Amaral, L.~Glass, J.~M. Hausdorff, P.~C. Ivanov, R.~G. Mark, J.~E. Mietus, G.~B. Moody, {Chung-Kang} Peng, and H.~E. Stanley.
\newblock Physiobank, physiotoolkit, and physionet: Components of a new research resource for complex physiologic signals.
\newblock \emph{Circulation}, 101\penalty0 (23):\penalty0 e215--e220, 2000.
\newblock \href{https://doi.org/10.1161/01.cir.101.23.e215}{\ttfamily\path{ doi:10.1161/01.cir.101.23.e215}}.

\bibitem[Harutyunyan et~al.(2019)Harutyunyan, Khachatrian, Kale, Steeg, and Galstyan]{HARUTYUNYANMultitaskLearningAndBenchmarkingWithClinicalTimeSeriesData}
H.~Harutyunyan, H.~Khachatrian, D.~C. Kale, G.~V. Steeg, and A.~Galstyan.
\newblock Multitask learning and benchmarking with clinical time series data.
\newblock \emph{Scientific Data}, 6\penalty0 (96), 6 2019.
\newblock \href{https://doi.org/10.1038/s41597-019-0103-9}{\ttfamily\path{ doi:10.1038/s41597-019-0103-9}}.

\bibitem[Henry et~al.(2016)Henry, Pylypchuk, Searcy, and Patel]{AdoptionOfElectronicHealthRecordSystemsAmongUSNonFederalAcuteCareHospitals}
J.~Henry, Y.~Pylypchuk, T.~Searcy, and V.~Patel.
\newblock Adoption of electronic health record systems among us non-federal acute care hospitals: 2008-2015.
\newblock \emph{ONC data brief}, 35, 05 2016.

\bibitem[Johnson et~al.(2016{\natexlab{a}})Johnson, Pollard, and Mark]{MIMIC3}
A.~E.~W. Johnson, T.~J. Pollard, and R.~G. Mark.
\newblock Mimic-iii clinical database (version 1.4).
\newblock 2016{\natexlab{a}}.
\newblock \href{https://doi.org/10.13026/c2xw26}{\ttfamily\path{ doi:10.13026/c2xw26}}.

\bibitem[Johnson et~al.(2016{\natexlab{b}})Johnson, Pollard, Shen, Lehman, Feng, Ghassemi, Moody, Szolovits, Celi, and Mark]{MIMICIIIAFreelyAccessibleCriticalCareDatabase}
A.~E.~W. Johnson, T.~J. Pollard, Lu~Shen, {Li-Wei}~H. Lehman, Mengling Feng, M.~Ghassemi, B.~Moody, P.~Szolovits, L.~A. Celi, and R.~G. Mark.
\newblock Mimic-iii, a freely accessible critical care database.
\newblock \emph{Scientific Data}, 3\penalty0 (160035), 05 2016{\natexlab{b}}.
\newblock \href{https://doi.org/10.1038/sdata.2016.35}{\ttfamily\path{ doi:10.1038/sdata.2016.35}}.

\bibitem[Jospin et~al.(2022)Jospin, Laga, Boussaid, Buntine, and Bennamoun]{HandsOnBayesianNeuralNetworksATutorialForDeepLearningUsers}
L.~V. Jospin, H.~Laga, F.~Boussaid, W.~Buntine, and M.~Bennamoun.
\newblock Hands-on bayesian neural networks—a tutorial for deep learning users.
\newblock \emph{IEEE Computational Intelligence Magazine}, 17\penalty0 (2):\penalty0 29--48, 2022.
\newblock \href{https://doi.org/10.1109/mci.2022.3155327}{\ttfamily\path{ doi:10.1109/mci.2022.3155327}}.

\bibitem[Kingma and Ba(2017)]{AdamAMethodForStochasticOptimization}
D.~P. Kingma and Jimmy Ba.
\newblock Adam: A method for stochastic optimization.
\newblock 2017.
\newblock \href{http://arxiv.org/abs/1412.6980}{{\ttfamily arXiv:1412.6980 [cs.LG]}}.

\bibitem[Kivlichan et~al.(2021)Kivlichan, Lin, Liu, and Vasserman]{MeasuringAndImprovingModelModeratorCollaborationUsingUncertaintyEstimation}
I.~D. Kivlichan, Zi~Lin, Jeremiah Liu, and L.~Vasserman.
\newblock Measuring and improving model-moderator collaboration using uncertainty estimation.
\newblock 2021.
\newblock \href{http://arxiv.org/abs/2107.04212}{{\ttfamily arXiv:2107.04212 [cs.LG]}}.

\bibitem[Kompa et~al.(2021)Kompa, Snoek, and Beam]{SecondOpinionNeededCommunicatingUncertaintyInMedicalMachineLearning}
B.~Kompa, J.~Snoek, and A.~L. Beam.
\newblock Second opinion needed: Communicating uncertainty in medical machine learning.
\newblock \emph{NPJ Digital Medicine}, 4\penalty0 (1):\penalty0 4, 1 2021.
\newblock \href{https://doi.org/10.1038/s41746-020-00367-3}{\ttfamily\path{ doi:10.1038/s41746-020-00367-3}}.

\bibitem[Kull et~al.(2019)Kull, Nieto, K\"{a}ngsepp, Filho, Song, and Flach]{BeyondTemperatureScalingObtainingWellCalibratedMultiClassProbabilitiesWithDirichletCalibration}
M.~Kull, M.~P. Nieto, M.~K\"{a}ngsepp, T.~S. Filho, Hao Song, and P.~Flach.
\newblock Beyond temperature scaling: Obtaining well-calibrated multi-class probabilities with dirichlet calibration.
\newblock In \emph{Advances in Neural Information Processing Systems}, volume~32, 2019.

\bibitem[Lakshminarayanan et~al.(2017)Lakshminarayanan, Pritzel, and Blundell]{SimpleAndScalablePredictiveUncertaintyEstimationUsingDeepEnsembles}
B.~Lakshminarayanan, A.~Pritzel, and C.~Blundell.
\newblock Simple and scalable predictive uncertainty estimation using deep ensembles.
\newblock In \emph{Advances in Neural Information Processing Systems}, volume~30, 2017.

\bibitem[Ledley and Lusted(1959)]{ReasoningFoundationsOfMedicalDiagnosis1959}
R.~S. Ledley and L.~B. Lusted.
\newblock Reasoning foundations of medical diagnosis: Symbolic logic, probability, and value theory aid our understanding of how physicians reason.
\newblock \emph{Science}, 130\penalty0 (3366):\penalty0 9--21, 07 1959.
\newblock \href{https://doi.org/10.1126/science.130.3366.9}{\ttfamily\path{ doi:10.1126/science.130.3366.9}}.

\bibitem[Li et~al.(2020)Li, Rao, Solares, Hassaine, Ramakrishnan, Canoy, Zhu, Rahimi, and Salimi-Khorshidi]{BEHRTTransformerforElectronic}
Yikuan Li, Shishir Rao, J.~R.~A. Solares, A.~Hassaine, R.~Ramakrishnan, D.~Canoy, Yajie Zhu, K.~Rahimi, and G.~Salimi-Khorshidi.
\newblock Behrt: Transformer for electronic health records.
\newblock \emph{Scientific Reports}, 10\penalty0 (7155), 04 2020.
\newblock \href{https://doi.org/10.1038/s41598-020-62922-y}{\ttfamily\path{ doi:10.1038/s41598-020-62922-y}}.

\bibitem[Li et~al.(2021)Li, Rao, Hassaine, R., Canoy, {Salimi-Khorshidi}, Mamouei, Lukasiewicz, and Rahimi]{DeepBayesianGaussianProcessesForUncertaintyEstimationInElectronicHealthRecords}
Yikuan Li, Shishir Rao, A.~Hassaine, Ramakrishnan R., D.~Canoy, G.~{Salimi-Khorshidi}, M.~Mamouei, T.~Lukasiewicz, and K.~Rahimi.
\newblock Deep bayesian gaussian processes for uncertainty estimation in electronic health records.
\newblock \emph{Scientific reports}, 11\penalty0 (20685), 10 2021.
\newblock \href{https://doi.org/10.1038/s41598-021-00144-6}{\ttfamily\path{ doi:10.1038/s41598-021-00144-6}}.

\bibitem[Liang et~al.(2021)Liang, Li, Zhang, Shen, Xu, Zheng, Wang, Tang, Lei, and Zhang]{AdoptionOfElectronicHealthRecordsEHRsInChinaDuringThePast10YearsConsecutiveSurveyDataAnalysisAndComparisonOfSinoAmericanChallengesAndExperiences}
Jun Liang, Ying Li, Zhang Zhang, Dongxia Shen, Jie Xu, Xu~Zheng, Tong Wang, Buzhou Tang, Jianbo Lei, and Jiajie Zhang.
\newblock Adoption of electronic health records (ehrs) in china during the past 10 years: Consecutive survey data analysis and comparison of sino-american challenges and experiences.
\newblock \emph{Journal of Medical Internet Research}, 23\penalty0 (2):\penalty0 e24813, 02 2021.
\newblock \href{https://doi.org/10.2196/24813}{\ttfamily\path{ doi:10.2196/24813}}.

\bibitem[Lim et~al.(2019)Lim, Lee, Hsu, and Wong]{BuildingTrustInDeepLearningSystemTowardsAutomatedDiseaseDetection}
Zhan~Wei Lim, Mong~Li Lee, Wynne Hsu, and Tien~Yin Wong.
\newblock Building trust in deep learning system towards automated disease detection.
\newblock \emph{Proceedings of the AAAI Conference on Artificial Intelligence}, 33\penalty0 (01):\penalty0 9516--9521, 07 2019.
\newblock \href{https://doi.org/10.1609/aaai.v33i01.33019516}{\ttfamily\path{ doi:10.1609/aaai.v33i01.33019516}}.

\bibitem[Liu et~al.(2020)Liu, Lin, Padhy, Tran, {Bedrax Weiss}, and Lakshminarayanan]{SimpleAndPrincipledUncertaintyEstimationWithDeterministicDeepLearningViaDistanceAwareness}
Jeremiah Liu, Zi~Lin, S.~Padhy, Dustin Tran, T.~{Bedrax Weiss}, and B.~Lakshminarayanan.
\newblock Simple and principled uncertainty estimation with deterministic deep learning via distance awareness.
\newblock In \emph{Advances in Neural Information Processing Systems}, volume~33, pages 7498--7512, 2020.

\bibitem[MacEachern and Forkert(2021)]{MachineLearningForPrecisionMedicine}
S.~J. MacEachern and N.~D. Forkert.
\newblock Machine learning for precision medicine.
\newblock \emph{Genome}, 64\penalty0 (4):\penalty0 416--425, 2021.
\newblock \href{https://doi.org/10.1139/gen-2020-0131}{\ttfamily\path{ doi:10.1139/gen-2020-0131}}.

\bibitem[MacKay(1992)]{APracticalBayesianFrameworkForBackpropagationNetworks}
D.~J.~C. MacKay.
\newblock {A Practical Bayesian Framework for Backpropagation Networks}.
\newblock \emph{Neural Computation}, 4\penalty0 (3):\penalty0 448--472, 05 1992.
\newblock \href{https://doi.org/10.1162/neco.1992.4.3.448}{\ttfamily\path{ doi:10.1162/neco.1992.4.3.448}}.

\bibitem[Menachemi and Collum(2011)]{BenefitsAndDrawbacksOfElectronicHealthRecordSystems}
N.~Menachemi and T.~H. Collum.
\newblock Benefits and drawbacks of electronic health record systems.
\newblock \emph{Risk Management and Healthcare Policy}, 4:\penalty0 47--55, 2011.
\newblock \href{https://doi.org/10.2147/rmhp.s12985}{\ttfamily\path{ doi:10.2147/rmhp.s12985}}.

\bibitem[Mullenbach et~al.(2018)Mullenbach, Wiegreffe, Duke, Sun, and Eisenstein]{ExplainablePredictionOfMedicalCodesFromClinicalText}
J.~Mullenbach, S.~Wiegreffe, J.~Duke, Jimeng Sun, and J.~Eisenstein.
\newblock Explainable prediction of medical codes from clinical text.
\newblock In \emph{Proceedings of the 2018 Conference of the North American Chapter of the Association for Computational Linguistics: Human Language Technologies}, volume~1, pages 1101--1111, 06 2018.
\newblock \href{https://doi.org/10.18653/v1/n18-1100}{\ttfamily\path{ doi:10.18653/v1/n18-1100}}.

\bibitem[Nazarovs et~al.(2021)Nazarovs, Mehta, Lokhande, and Singh]{GraphReparameterizationsForEnabling1000MonteCarloIterationsInBayesianDeepNeuralNetworks}
J.~Nazarovs, R.~R. Mehta, V.~S. Lokhande, and V.~Singh.
\newblock Graph reparameterizations for enabling 1000+ monte carlo iterations in bayesian deep neural networks.
\newblock In \emph{Proceedings of the Thirty-Seventh Conference on Uncertainty in Artificial Intelligence}, volume 161, pages 118--128, 07 2021.

\bibitem[Neal(1996)]{BayesianLearningForNeuralNetworks}
R.~M. Neal.
\newblock \emph{Bayesian Learning For Neural Networks}.
\newblock Springer New York, 1996.
\newblock \href{https://doi.org/10.1007/978-1-4612-0745-0}{\ttfamily\path{ doi:10.1007/978-1-4612-0745-0}}.

\bibitem[Pang et~al.(2021)Pang, Jiang, Kalluri, Spotnitz, Chen, Perotte, and Natarajan]{CEHRBERTIncorporatingTemporalInformationFrom}
Chao Pang, Xinzhuo Jiang, K.~S. Kalluri, M.~Spotnitz, RuiJun Chen, A.~Perotte, and K.~Natarajan.
\newblock Cehr-bert: Incorporating temporal information from structured ehr data to improve prediction tasks.
\newblock In \emph{Proceedings of Machine Learning for Health}, volume 158, pages 239--260, 12 2021.

\bibitem[Pearce et~al.(2020)Pearce, Leibfried, and Brintrup]{UncertaintyInNeuralNetworksApproximatelyBayesianEnsembling}
T.~Pearce, F.~Leibfried, and A.~Brintrup.
\newblock Uncertainty in neural networks: Approximately bayesian ensembling.
\newblock In \emph{Proceedings of the Twenty Third International Conference on Artificial Intelligence and Statistics}, volume 108, pages 234--244, 08 2020.

\bibitem[Peng et~al.(2019)Peng, Long, Shen, Wang, Jiang, and Blumenstein]{TemporalSelfAttentionNetworkforMedicalConceptEmbedding}
Xueping Peng, Guodong Long, Tao Shen, Sen Wang, Jing Jiang, and M.~Blumenstein.
\newblock Temporal self-attention network for medical concept embedding.
\newblock In \emph{2019 IEEE International Conference on Data Mining}, pages 498--507, 11 2019.
\newblock \href{https://doi.org/10.1109/icdm.2019.00060}{\ttfamily\path{ doi:10.1109/icdm.2019.00060}}.

\bibitem[Qiu(2020)]{ModelingUncertaintyInDeepLearningModelsOfElectronicHealthRecords}
Riyi Qiu.
\newblock \emph{Modeling Uncertainty in Deep Learning Models of Electronic Health Records}.
\newblock PhD thesis, The University of North Carolina at Charlotte, 2020.

\bibitem[Qiu et~al.(2019)Qiu, Jia, Hadzikadic, Dulin, Niu, and Wang]{ModelingTheUncertaintyInElectronicHealthRecordsABayesianDeepLearningApproach}
Riyi Qiu, Yugang Jia, M.~Hadzikadic, M.~Dulin, Xi~Niu, and Xin Wang.
\newblock Modeling the uncertainty in electronic health records: A bayesian deep learning approach.
\newblock 2019.
\newblock \href{http://arxiv.org/abs/1907.06162}{{\ttfamily arXiv:1907.06162 [cs.LG]}}.

\bibitem[Raghupathi and Raghupathi(2014)]{BigDataAnalyticsInHealthcarePromiseAndPotential}
W.~Raghupathi and V.~Raghupathi.
\newblock Big data analytics in healthcare: Promise and potential.
\newblock \emph{Health Information Science and Systems}, 2\penalty0 (3), 2014.
\newblock \href{https://doi.org/10.1186/2047-2501-2-3}{\ttfamily\path{ doi:10.1186/2047-2501-2-3}}.

\bibitem[Rasmy et~al.(2021)Rasmy, Xiang, Xie, Tao, and Zhi]{MedBERTPretrainedContextualizedEmbeddingsOnLargescaleStructuredElectronicHealthRecordsForDiseasePrediction}
L.~Rasmy, Yang Xiang, Ziqian Xie, Cui Tao, and Degui Zhi.
\newblock Med-bert: Pretrained contextualized embeddings on large-scale structured electronic health records for disease prediction.
\newblock \emph{NPJ Digital Medicine}, 4\penalty0 (86), 05 2021.
\newblock \href{https://doi.org/10.1038/s41746-021-00455-y}{\ttfamily\path{ doi:10.1038/s41746-021-00455-y}}.

\bibitem[Ristevski and Chen(2018)]{BigDataAnalyticsInMedicineAndHealthcare}
B.~Ristevski and Ming Chen.
\newblock Big data analytics in medicine and healthcare.
\newblock \emph{Journal of Integrative Bioinformatics}, 15\penalty0 (3):\penalty0 20170030, 2018.
\newblock \href{https://doi.org/10.1515/jib-2017-0030}{\ttfamily\path{ doi:10.1515/jib-2017-0030}}.

\bibitem[Ruder(2017)]{AnOverviewOfGradientDescentOptimizationAlgorithms}
S.~Ruder.
\newblock An overview of gradient descent optimization algorithms.
\newblock 2017.
\newblock \href{http://arxiv.org/abs/1609.04747}{{\ttfamily arXiv:1609.04747 [cs.LG]}}.

\bibitem[Shang et~al.(2019)Shang, Ma, Xiao, and Sun]{PreTrainingOfGraphAugmentedTransformersForMedicationRecommendation}
Junyuan Shang, Tangfei Ma, Cao Xiao, and Jimeng Sun.
\newblock Pre-training of graph augmented transformers for medication recommendation.
\newblock In \emph{Proceedings of the Twenty-Eighth International Joint Conference on Artificial Intelligence}, pages 5953--5959, 07 2019.
\newblock \href{https://doi.org/10.24963/ijcai.2019/825}{\ttfamily\path{ doi:10.24963/ijcai.2019/825}}.

\bibitem[Shickel et~al.(2018)Shickel, Tighe, Bihorac, and Rashidi]{DeepEHRASurveyOfRecentAdvancesInDeepLearningTechniquesForElectronicHealthRecordEHRAnalysis}
B.~Shickel, P.J. Tighe, A.~Bihorac, and P.~Rashidi.
\newblock Deep ehr: A survey of recent advances in deep learning techniques for electronic health record {(EHR)} analysis.
\newblock \emph{IEEE Journal of Biomedical and Health Informatics}, 22\penalty0 (5):\penalty0 1589--1604, 09 2018.
\newblock \href{https://doi.org/10.1109/jbhi.2017.2767063}{\ttfamily\path{ doi:10.1109/jbhi.2017.2767063}}.

\bibitem[Si et~al.(2021)Si, Du, Li, Jiang, Miller, Wang, {Jim Zheng}, and Roberts]{DeepRepresentationLearningOfPatientDataFromElectronicHealthRecordsEHRASystematicReview}
Yuqi Si, Jingcheng Du, Zhao Li, Xiaoqian Jiang, T.~Miller, Fei Wang, W.~{Jim Zheng}, and K.~Roberts.
\newblock Deep representation learning of patient data from electronic health records (ehr): A systematic review.
\newblock \emph{Journal of Biomedical Informatics}, 115\penalty0 (103671), 03 2021.
\newblock \href{https://doi.org/10.1016/j.jbi.2020.103671}{\ttfamily\path{ doi:10.1016/j.jbi.2020.103671}}.

\bibitem[Smith(2013)]{UncertaintyQuantificationTheoryImplementationAndApplications}
R.~C. Smith.
\newblock \emph{Uncertainty Quantification: Theory, Implementation, and Applications}.
\newblock Society for Industrial and Applied Mathematics, Philadelphia, PA, 2013.
\newblock \href{https://doi.org/10.1137/1.9781611973228}{\ttfamily\path{ doi:10.1137/1.9781611973228}}.

\bibitem[Song et~al.(2017)Song, Rajan, Thiagarajan, and Spanias]{AttendAndDiagnoseClinicalTimeSeriesAnalysisUsingAttentionModels}
Huan Song, D.~Rajan, J.~J. Thiagarajan, and A.~Spanias.
\newblock Attend and diagnose: Clinical time series analysis using attention models.
\newblock 2017.
\newblock \href{http://arxiv.org/abs/1711.03905}{{\ttfamily arXiv:1711.03905 [stat.ML]}}.

\bibitem[Sullivan(2015)]{IntroductionToUncertaintyQuantification}
T.~J. Sullivan.
\newblock \emph{Measures of Information and Uncertainty}, pages 75--90.
\newblock Springer International Publishing, Cham, 2015.
\newblock \href{https://doi.org/10.1007/978-3-319-23395-6-5}{\ttfamily\path{ doi:10.1007/978-3-319-23395-6-5}}.

\bibitem[Tonekaboni et~al.(2019)Tonekaboni, Joshi, McCradden, and Goldenberg]{WhatCliniciansWantContextualizingExplainableMachineLearningForClinicalEndUse}
S.~Tonekaboni, S.~Joshi, M.~D. McCradden, and A.~Goldenberg.
\newblock What clinicians want: Contextualizing explainable machine learning for clinical end use.
\newblock In \emph{Proceedings of the 4th Machine Learning for Healthcare Conference}, volume 106, pages 359--380, 08 2019.

\bibitem[Vaswani et~al.(2017)Vaswani, Shazeer, Parmar, Uszkoreit, Jones, Gomez, Kaiser, and Polosukhin]{AttentionIsAllYouNeed}
A.~Vaswani, N.~Shazeer, N.~Parmar, J.~Uszkoreit, L.~Jones, A.~N. Gomez, \L. Kaiser, and I.~Polosukhin.
\newblock Attention is all you need.
\newblock In \emph{Advances in Neural Information Processing Systems}, volume~30, 2017.

\bibitem[Wang et~al.(2019)Wang, Xu, Jin, Li, Xie, and Wang]{Inpatient2VecMedicalRepresentationLearningForInpatients}
Ying Wang, Xiao Xu, Tao Jin, Xiang Li, Guotong Xie, and Jianmin Wang.
\newblock Inpatient2vec: Medical representation learning for inpatients.
\newblock In \emph{2019 IEEE International Conference on Bioinformatics and Biomedicine}, pages 1113--1117, 11 2019.
\newblock \href{https://doi.org/10.1109/bibm47256.2019.8983281}{\ttfamily\path{ doi:10.1109/bibm47256.2019.8983281}}.

\bibitem[Wenzel et~al.(2020)Wenzel, Snoek, Tran, and Jenatton]{HyperparameterEnsemblesForRobustnessAndUncertaintyQuantification}
F.~Wenzel, J.~Snoek, Dustin Tran, and R.~Jenatton.
\newblock Hyperparameter ensembles for robustness and uncertainty quantification.
\newblock In \emph{Advances in Neural Information Processing Systems}, volume~33, pages 6514--6527, 2020.

\bibitem[Wu et~al.(2017)Wu, Cheng, Kaddi, Venugopalan, Hoffman, and Wang]{OmicAndElectronicHealthRecordBigDataAnalyticsForPrecisionMedicine}
{Po-Yen} Wu, {Chih-Wen} Cheng, C.~D. Kaddi, J.~Venugopalan, R.~Hoffman, and May~Dongmei Wang.
\newblock –omic and electronic health record big data analytics for precision medicine.
\newblock \emph{IEEE Transactions on Biomedical Engineering}, 64\penalty0 (2):\penalty0 263--273, 2017.
\newblock \href{https://doi.org/10.1109/tbme.2016.2573285}{\ttfamily\path{ doi:10.1109/tbme.2016.2573285}}.

\end{thebibliography}

\newpage
\appendix
\section{Appendix}

\subsection{ELBO Loss KL Term} \label{sec:appx:KLTermderiv}

The radial distribution utilized as prior for the BNN models in this work is defined by: 

\begin{equation} \label{eq:radial}
\begin{aligned}
w_{radial} &= \mu_q + \sigma_q \frac{\epsilon}{||\epsilon||} r \qquad r = N(0,1) \qquad \epsilon = N(0,1)
\end{aligned}
\end{equation}

No closed form solution exists for the KL term but \citep{RadialBayesianNeuralNetworksBeyondDiscreteSupportInLargeScaleBayesianDeepLearning} derived a Monte Carlo approximation. \citep{GraphReparameterizationsForEnabling1000MonteCarloIterationsInBayesianDeepNeuralNetworks} showed that for selected combinations of distributions the computation can be further simplified. This results in a computational graph (in modern deep learning frameworks) independent of the number of samples, leading to vastly improved performance and decreased memory demands. Applying both modifications equation \ref{eq:c1c2} shows the final form of the KL term.

\begin{equation} \label{eq:c1c2}
\begin{aligned}
KL[q(w|\theta_q)||p(w)] &= \int q(w|\theta_q) \log(q(w|\theta_q))dw - \int q(w|\theta_q) \log(p(w))dw \\
&=\mathcal{L}_{entropy} - \mathcal{L}_{crossentropy} \\
\mathcal{L}_{entropy} &= \log(\sigma_q) \\
\mathcal{L}_{crossentropy} &= -0.5\log(\sigma_p^2\sqrt{2\pi}) \\
&-\frac{1}{2\sigma_p^2}
(\mu_{q-p}^2+2\mu_{q-p}\sigma_{q}\mathbb{E}_{\epsilon}\Bigl[\frac{\epsilon}{||\epsilon||} r\Bigr]+\sigma_{q}^2\mathbb{E}_{\epsilon}\Bigl[(\frac{\epsilon}{||\epsilon||} r)^2\Bigr])
\end{aligned}
\end{equation}

\subsection{ECE-Loss Regularisation} \label{sec:appx:eceregularistaion}

We found the ECE loss with fixed bins to destabilize the training procedure for the ensemble model while showing no such signs for the Bayesian model. By randomising the bins and averaging over binnings $\boldsymbol{K}$ we managed to stabilize training successfully (see equation \ref{eq:eceloss_reg}). Randomization was achieved by drawing the required bin edges from a uniform distribution $U(0+\epsilon, 1-\epsilon)$ and sorting the resulting set including the limits 0 and 1. Thus we can generate random binnings $\boldsymbol{K}$ by sampling from this generator denoted as $\mathcal{G}_{\boldsymbol{K}}$. The number of samples used for calculation is denoted by $G$.

\begin{align} \label{eq:eceloss_reg}
\textit{ECE-Loss} = \frac{1}{S*M*G}\sum_{z\in s}\sum_{m\in \boldsymbol{M}}\sum_{\boldsymbol{K}\sim \mathcal{G}_{\boldsymbol{K}}}\sum_{k\in \boldsymbol{K}} conf_{k,m} - acc_{k,m}
\end{align}

We add a normal Cross Entropy loss to the ECE loss. This is to prevent the trivial solution by zeroing $\boldsymbol{W}^{MC}$ and setting $\boldsymbol{b}^{MC}$ to predict the probabilities of classes inherent to the data set, achieving theoretically perfect ECE loss \citep{BeyondTemperatureScalingObtainingWellCalibratedMultiClassProbabilitiesWithDirichletCalibration}. The combination ensures that calibration is not done at the expense of the correct answer.

\subsection{Model Selection and Configurations} \label{sec:appx:modelselectionandconfigurations}

As the goal of the work was not to optimize performance but rather to research the adequacy of the selected methods for detecting EU, we ran a small parameter search over three structurally significant parameters for each model to at least match the performance of comparable models. An overview of the search space is shown in table \ref{tab:paramsearch}. 

\begin{table}[H]
\centering
\begin{tabular}{|l t c|c|c|}
\hline
 & embedding size $(D)$ & \# heads (Multi Head Attention) & \# layers \\
\hhline{|= t =|=|=|}
Search Space & [32, 64, 128] & [2, 4] & [1, 2, 3] \\
\hhline{|= t =|=|=|}
BT & 64 & 2 & 1 \\
\hline
TE (33 models) 
& \makecell[c]{30\%\hspace{10pt}32\\36\%\hspace{10pt}64\\33\%\hspace{10pt}128} 
& \makecell[c]{49\%\hspace{10pt}2\\51\%\hspace{10pt}4}
& \makecell[c]{36\%\hspace{10pt}1\\36\%\hspace{10pt}2\\27\%\hspace{10pt}3}  \\
\hline
\end{tabular}
\caption{Summary of the parameter search conducted for both model types and selected model configurations. For the TE the distribution for each parameter is given.}
\label{tab:paramsearch}
\end{table}

Due to the reduced complexity and size of the data set, training seemed to favour smaller models which converged much quicker for the BT model. The best configuration was selected based on the averaged AUC ROC scores over all cross validation splits. For the TE model, the best 33 models were selected based on the same procedure. The models varied with no clear indication in favour of any particular configuration in their parametrisation. The configurations for both model types are shown in table \ref{tab:paramsearch}.
	
\end{document}